\documentclass[10pt,a4paper]{article}
\usepackage{amsmath,graphicx,times}
\usepackage[textwidth=12.75cm]{geometry}
\usepackage{color}

\newcommand{\dif}{\textrm{d}}
\newcommand{\E}{\mathbb{E}}

\usepackage{amsfonts}
\usepackage{algorithm,algorithmic}
\usepackage{amsmath}
\usepackage{xfrac}

\DeclareMathOperator*{\argmax}{argmax }
\DeclareMathOperator*{\argmin}{argmin }

\usepackage{graphicx}

\usepackage{subcaption}

\usepackage{charter}

\usepackage{fancyhdr}
\usepackage[latin1]{inputenc}

\begin{document}

\title{
	%	\setlength{\parindent}{0pt}
	%	\setlength{\parskip}{0pt}
	%	\vspace*{\stretch{1}}
	\rule{\linewidth}{1pt}
	\vspace{5pt}
	\textbf{Efficient Learning of Harmonic Priors for Pitch Detection in Polyphonic Music}
	\rule{\linewidth}{1pt}
	%	\vspace*{\stretch{2}}
}

\author{\textbf{Pablo A. Alvarado}, \hspace{2.5pt} \textbf{Dan Stowell} \\ Centre for Digital Music \\ Queen Mary University of London}
\date{}
\maketitle

\begin{abstract}
\noindent
%% Background
Automatic music transcription (AMT) aims to infer a \textit{latent} symbolic representation of a piece of music (piano-roll), given a corresponding \textit{observed} audio recording. Transcribing polyphonic music (when multiple notes are played simultaneously) is a challenging problem, due to highly structured overlapping between harmonics.
%
%New Method
%We seek to develop audio content analysis Bayesian models that unify prior knowledge of underlying physical mechanisms and music-driven rules that govern the nature of acoustic signals. 
%
We study whether the introduction of physically inspired Gaussian process (GP) priors  into audio content analysis models improves the extraction of patterns required for AMT.
Audio signals are described as a linear combination of sources. Each source is decomposed into the product of an amplitude-envelope, and a quasi-periodic \textit{component} process.
We introduce the Mat\'{e}rn spectral mixture (MSM) kernel for describing frequency content of singles notes.
We consider two different regression approaches. In the \textit{sigmoid} model every pitch-activation is independently non-linear transformed. In the \textit{softmax} model 
several activation GPs are jointly non-linearly transformed. This introduce cross-correlation between activations.
We use variational Bayes for approximate inference.
%Results
We empirically evaluate how these models work in practice transcribing polyphonic music.
%Comparison with Existing Method(s)
%
We demonstrate that rather than encourage dependency between activations, what is relevant for improving pitch detection is to learnt priors that fit the frequency content of the sound events to detect. 
The Python code  is available at \texttt{https://github.com/PabloAlvarado/MSMK}.
\end{abstract}

%\begin{keywords}
%Gaussian processes, spectral mixture kernel, pitch detection, automatic music transcription.
%\end{keywords}

% PAPER SECTIONS
\section{Introduction} 
\label{sec:intro}
In music information research, the aim of audio content analysis is to infer musical concepts (pitch, melody, chords, onset, beat, tempo, rhythm) which are present but hidden in the audio data \cite{serra13}. 
Then, perhaps the most general application is recovering the score (symbolic representation) of a music track given only the audio recording \cite{muller11}. This is known as automatic music transcription (AMT) \cite{benetos13}.
Transcribing polyphonic music (when multiple notes are played simultaneously) is a challenging problem, especially in its more unconstrained form when the task is performed on an arbitrary acoustical input \cite{benetos12}.
This is because simultaneous notes cause a highly structured overlap of harmonics in the acoustic signal \cite{Sigtia16}.
Besides,
a single note produced by a music instrument is not simply a fixed-duration sine wave at an appropriate frequency, but rather a full spectrum of harmonics with an attack and  decay in intensity. These spectrum evolution is instrument dependent, and therefore must be learned in a recording-specific way \cite{Kirkpatrick14, Cheng16}.
The polyphony together with complex harmonic structure of sound events creates a source-separation problem at the heart of the transcription task \cite{Kirkpatrick14}.

We seek to take advantage of the strong statistical structure that music acoustic signals have \cite{cemgil10}.
Then, the goal of this research is to develop audio content analysis Bayesian models that naturally bring together prior knowledge about the underlying physical mechanisms and rules based on music theory that govern the global nature of acoustic signals.  
We study whether the introduction of several musically-physically driven patterns in the prior of probabilistic models improves the analysis, which in turn can increase the quality of automatic transcription. Our approach has at the core Gaussian processes (GPs). 
GPs have been extensively used for modelling audio recordings. 
GPs are used to consider time-frequency analysis as probabilistic inference \cite{Turner14}, source separation \cite{Liutkus11,Yoshii13b, Adam16}, 
and for estimating spectral envelope and fundamental frequency of a speech signal \cite{Yoshii13}. GPs  for music genre classification and emotion estimation were investigated in \cite{Markov14}. 
Finally, in \cite{Ohishi14} a mixture of Gaussian process experts was used for predicting sung melodic contour with expressive dynamic fluctuations.

%In the first stage report of the PhD research we focused on designing priors that capture physical or mechanistic high-level properties of the analysed audio signals, specifically: non-stationarity, time dynamics (local periodicity, and non constant amplitude envelope), and spectral harmonic content. The outline of this model is described in section \ref{sec:model_variant_one}. Results obtained with this approach were presented at the IEEE International Workshop on Machine Learning for Signal Processing (MLSP 2016) \cite{alvarado16}.

%The contribution in this second report  are: 
%
Similar to \cite{Adams08}, we propose a model where two GPs are multiplied. Here several GPs are jointly non-linear transformed using the \textit{softmax} function. We call this the softmax model. This comes as a principled way to introduce correlation between pitch activations, encouraging them to reflect two properties: non-negativity, and sparsity; to enable few pitches to be active at certain time. 
Second, we introduce what we call the \textit{Mat\'{e}rn spectral mixture} (MSM) kernel, in order to unify in one single covariance function the description of long-term dependences driven by rhythm on pitch activations, as well as the harmonic content of sound events. %Section \ref{sec:model_variant_two} contains additional preliminary work on the development of how to introduce more music structure in the kernel.
Analogous to \cite{Wilson13}, we model a spectral density as a mixture of basis functions, but for efficient inference instead of Gaussians, we use Lorentzian functions \cite{Kanevsky09}, following the Fourier transform of the Mat\'{e}rn-$\frac{1}{2}$ kernel \cite{Hensman16}. In this work, we use the Mat\'{e}rn spectral mixture covariance function to encourage the model prior to reflect the clearly evident complex harmonic content present in mixture signals which can be learnt in advance from of isolated sounds.
Third, we increase the model scalability through approximate methods using variational Bayes, enabling the analysis of audio signals with several seconds of duration.
Finally, in comparison with the model presented in \cite{alvarado16}, with the proposed approach the amount of model parameters becomes independent of the total sound events present in the audio recording. Moreover, to know \textit{a priori} the number of sound events becomes inessential, as this quantity is learn directly from audio.

The paper is organized as follows. Section \ref{sec:materials_methods} introduces the GPs model for pitch detection. Two different variants of the base model are presented in sections \ref{sec:sigmoid_model} and  \ref{sec:softmax_model}. In section \ref{sec:inference}, we provide details for learning in frequency domain the parameters of the MSM kernel. We empirically evaluated how the proposed framework works in practice transcribing polyphonic music recordings (section \ref{sec:amt_poly}). Final conclusions are given in section \ref{sec:conclusions}. %, or as a post-processing stage assuming we already have noisy observations of the activations (results section \ref{sec:Multi_GPs}). %Finally, the work plan for the rest of the PhD research is presented in section \ref{sec.workplan}. 
\section{Method}
\label{sec:materials_methods}
%
%
%\subsection{Gaussian processes}
Gaussian processes (GPs) are at the core of the modelling approach presented in this work.
GP-based machine learning is a powerful Bayesian paradigm for nonparametric nonlinear regression and classification \cite{Sarkka2013}. 
GPs can be defined as distributions over functions such that any finite number of function evaluations $\textbf{f}=[f(t_1),\cdots, f(t_N)]$, have a jointly normal distribution \cite{rasmussen05}. A GP is completely specified by its mean function $\mu(t) = \mathbb{E}[f(t)]$ ,
%
%\begin{align}
%\mu(t) = \mathbb{E}[f(t)],
%\end{align}
%
and its kernel or covariance  
\begin{align}\label{e.covgen}
k(t,t') = 
%\text{Cov}[f(t),f(t')]
\mathbb{E}\left[ (f(t) - \mu(t)) ( f(t') - \mu(t'))\right] ,
\end{align} 
where $k(t,t')$ has free hyper-parameters $\boldsymbol{\theta}$. 
We write the GP as
%
%\begin{align*}
$
f(t) \sim \mathcal{GP}(\mu(t),k(t,t')).
$
%\end{align*}
%
The form of \eqref{e.covgen} captures high-level properties of the unknown function $f(t)$, which in turn determines how the model generalizes or extrapolates \cite{lloyd2014}. We used GPs for modelling
%either the components $\left\lbrace w_m(t)\right\rbrace_{m=1}^{M} $ or 
both, amplitude-envelope and component functions.

\subsection{Gaussian process model for pitch detection}
Recall that automatic music transcription aims to infer a \textit{latent} symbolic representation, such as piano-roll or score, given an \textit{observed} audio recording. Piano-roll refers to a matrix representation of musical notes across time \cite{benetos13,cemgil10}.
From a Bayesian latent variable perspective \cite{Blei16}, transcription consists in updating our beliefs about the symbolic description for a certain piece of music, after observing a corresponding audio recording. 
%%%%
%%%This traduces to the computation of a conditional distribution over the piano-roll, that is
%%%%
%%%$
%%%%\begin{align}\label{e.model_intro}
%%%p(\text{piano-roll}|\text{signal}) = \frac{p(\text{signal}|\text{piano-roll}) \times p(\text{piano-roll})}{p(\text{signal})},
%%%%\end{align}
%%%$
%%%%
%%%where $p(\text{signal}|\text{piano-roll})$ represents how likely the observed signal is under an specific piano-roll, our prior beliefs about the symbolic representation are encoded in $p(\text{piano-roll})$, and $p(\text{signal})$ is the evidence. 
%%%%
As in \cite{Yoshii13}, we approach the transcription problem from a time-domain source separation perspective.
That is, given an audio recording $\mathcal{D}=\left\lbrace y_n, t_n \right\rbrace_{n=1}^{N}$, we seek to formulate a generative probabilistic model that describes how the observed polyphonic signal (mixture of sources) was generated and, moreover, that allows us to infer the latent variables associated with the piano-roll representation. 
To do so, we use the regression model 
%
%\begin{align}\label{e.general_f}
$
y_n = f(t_n) + \epsilon,
$
%\end{align}
%	
where $y_n$ is the value of the analysed polyphonic signal at time $t_n$, the noise follows a normal distribution $\epsilon \sim \mathcal{N}(0,\sigma^2)$, and the function $f(t)$ is a random process composed by a linear combination of $M$ \textit{sources} $\left\lbrace  f_m(t) \right\rbrace _{m=1}^{M} $.
%
%\begin{align}\label{equ:sources_sum}
%f(t) = \sum_{m=1}^{M} f_{m}(t).
%\end{align}
%		
Each source is decomposed into the product of two factors, an amplitude-envelope or activation function $\phi_m(t)$, and a quasi-periodic or component function $w_{m}(t)$. The overall model is then 
\begin{align}\label{equ:factors_sum}
y(t) =  \sum_{m=1}^{M} \phi_{m}(t) w_{m}(t) + \epsilon.
\end{align}
%		
%% MATHEMATICAL MODEL 
%We can interpret the set $\left\lbrace w_{m}(t)\right\rbrace_{m=1}^{M}$ as a dictionary  where each component $ w_{m}(t)$ is a quasi-periodic stochastic function with a defined pitch. Likewise, each stochastic function in $\left\lbrace \phi_{m}(t)\right\rbrace_{m=1}^{M}$ represents the respectively time dependent activation of an specific pitch throughout the analysed piece for music, i.e. represent each row of the piano roll matrix. (be sure they are not negative).
%
%
%
We can interpret the set $\left\lbrace w_{m}(t)\right\rbrace_{m=1}^{M}$ as a dictionary  where each component $ w_{m}(t)$ is a quasi-periodic stochastic function with a defined pitch. 
Likewise, each stochastic function in $\left\lbrace \phi_{m}(t)\right\rbrace_{m=1}^{M}$ represents a row of the piano roll-matrix, i.e the time dependent non-negative activation of a specific pitch throughout the analysed piece of music.

%
%\begin{align}\label{equ:model_variant2}
%y(t) = \sum_{m=1}^{M} \phi_{m}(t) w_{m}(t) + \epsilon,
%\end{align}
%
Components $\left\lbrace w_{m}(t)\right\rbrace_{m=1}^{M}$ follow 
%
%\begin{align}
$
w_m(t) \sim \mathcal{GP}(0,k_m(t,t')),
$
%\end{align}
%
%
%where the covariance $k_m(t,t')$ reflect the frequency content of the $m^{\text{th}}$ component, and has the form of a MSM kernel, described shortly in section \ref{sec:msm_kernel}. 
where the covariance function $k_m(t,t')$ reflects the frequency content of the $m^{\text{th}}$ component, and has the form of a MSM kernel (section \ref{sec:msm_kernel}). 
In prior work \cite{alvarado16} only component functions followed GPs, whereas the amplitude-envelopes were deterministic functions. 
Here the flexibility of activations $\left\lbrace \phi_{m}(t)\right\rbrace_{m=1}^{M}$ increases by treating them as GPs non-linearly transformed either independently or jointly,  by using the sigmoid function (section \ref{sec:sigmoid_model}) or the softmax function (section \ref{sec:softmax_model}) respectively. 

\subsubsection{Sigmoid model}
\label{sec:sigmoid_model}
To guarantee the activations to be non-negative we apply non-linear transformations to GPs. To do so, we use the sigmoid function 
%
%\begin{align}\label{e:sigmoid}
$
\sigma(x) = \left[ 1 + \exp(-x) \right]^{-1},
$
%\end{align}
%
also applied for GP binary classification  \cite{rasmussen05}. The activations now are defined as 
$
%\begin{align*}
\phi_m(t) = \sigma( {g_{m}(t)} ),
%\end{align*}
$
where $ \left\lbrace g_{m}(t)\right\rbrace_{m=1}^{M} $ are GPs. The sigmoid model  follows
\begin{align}\label{equ:sigmoid_model}
y(t)=
\sum_{m = 1}^{M}
\sigma( {g_{m}(t)} )
w_{m}(t) 
+ \epsilon.
\end{align}
%
%This idea is illustrated with a simple example. we use \eqref{equ:sigmoid_model} for modelling the isolated sound event in a real bass recording shown in Fig. \ref{fig:real_and_approx_bass} (b) top. In this case there is one single pitch, i.e. $M=1$. The extracted activation $\phi(t)$ correspond to the green line in Fig. \ref{fig:comp_and_act_bass} top, whereas the inferred quasi-periodic component $w(t)$ corresponds to Fig. \ref{fig:comp_and_act_bass} bottom. The approximated signal is depicted in Fig. \ref{fig:real_and_approx_bass} (b) bottom. 
%

%\subsubsection{Single pitch approach}

%\subsubsection{"Leave one out" approach (In case there is time and space)}

\subsubsection{Softmax model}
\label{sec:softmax_model}
To enhance sparsity we use the \textit{softmax}, or \textit{normalized exponential} function for defining the form of the activations, that is
\begin{align}\label{e:softmax}
\phi_m(t) = \frac{\exp(g_{m}(t))}{\sum_{\forall j}  \exp(g_{j}(t))},
\end{align}
where $\left\lbrace g_{j}(t)\right\rbrace_{j=1}^{M}$ are GPs  \cite{murphy12, bishop06}. Similarly to the sigmoid function, the \textit{softmax} \eqref{e:softmax} enforces the activations to be non-negative as well as to be bounded between $0$ and $1$. Furthermore, \eqref{e:softmax} introduces dependences between all activations.
The sparsity is enhanced because
%The nice property of this function is that
%
$
%\begin{align*}
\sum_{\forall m} \phi_{m}(t) = 1,
%\end{align*}
$
for all $t$.
With this property we can encourage to activate only one or a few pitches at certain time; if the pitch $j$ explains better the audio signal at time $t_n$, then the activation $\phi_{j}(t_n) \approx 1$, therefore it follows that the other activations $\phi_{i}(t_n) \approx 0$ for all $i \neq j$. 
%
%We assume that the functions $g_{m}(t)$  in \eqref{e:softmax} follow GPs. 
%%
%\begin{align*}
%g_{m}(t) = \mathcal{GP}(0,\dot{k}_{m}(t,t')),
%\end{align*}
%
%where $\dot{k}_{m}(t,t')$ follows a 
% mMSM kernel (section \ref{sec:msm_kernel}) with low frequency periodicity related with rhythm.
%
The softmax model corresponds to 
\begin{align}\label{equ:softmax_model}
y(t)=
\frac{1}{\sum_{j = 0}^{M}  \exp(g_{j}(t))}
\sum_{m = 0}^{M}
%\left(  
\exp(g_{m}(t))
%\right) 
w_{m}(t) 
+ \epsilon,
\end{align}
where we choose the component process $w_{0}(t) = 0$ for all $t$ to allow for silence or rest. The activation $\phi_{0}(t)$ is equal to $1$ only when there is silence in the audio recording.
%
%As an example, for two pitches we have $M = 2$, that is 

\subsection{The Mat\'{e}rn spectral mixture kernel}

\label{sec:msm_kernel}
%CC Motivation

\begin{figure}[th!]
	\centering
	\begin{subfigure}[t]{0.5\columnwidth}
		\centering
		\includegraphics[width=1.0\columnwidth]{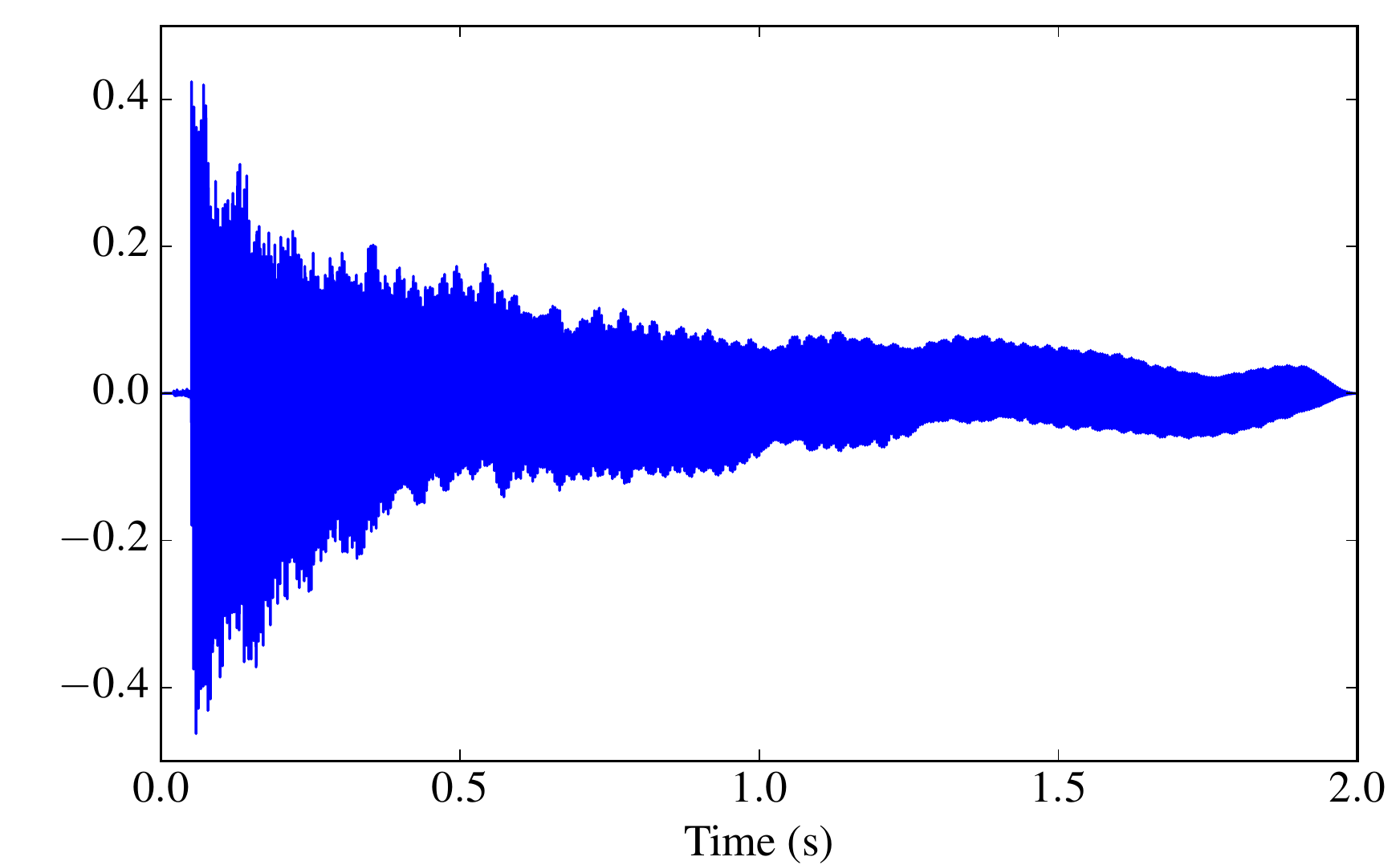}
		\caption{}
		\label{fig:waveform}
	\end{subfigure}%
	%\\
	\centering
	\begin{subfigure}[t]{0.5\columnwidth}
		\centering
		\includegraphics[width=1.0\columnwidth]{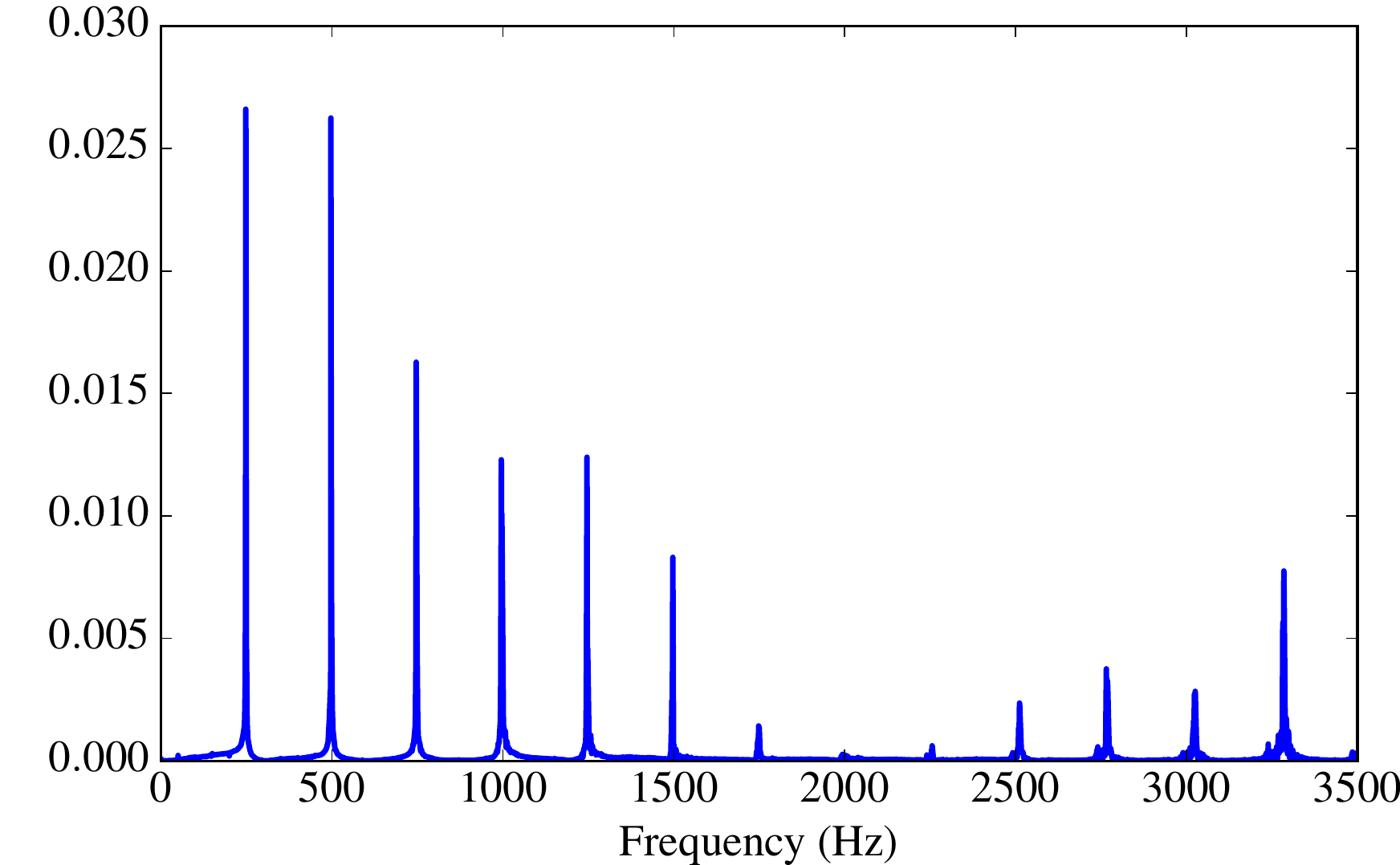}
		\caption{}
		\label{fig:FT_waveform}
	\end{subfigure}
	\\
	\centering
	\begin{subfigure}[t]{1.0\columnwidth}
		\centering
		\includegraphics[width=1.0\columnwidth]{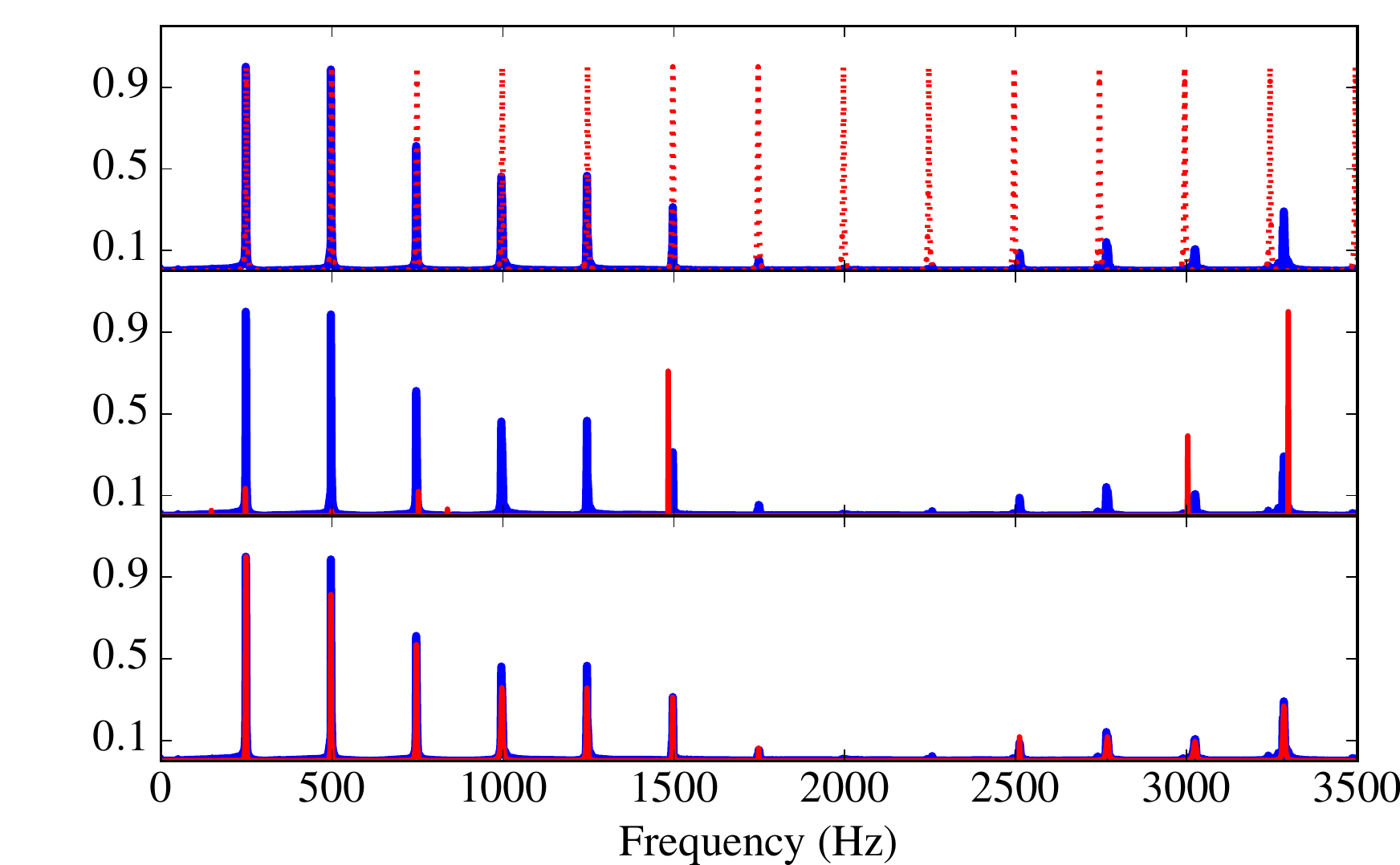}
		\caption{}
		\label{fig:compare_ML_TM_FL}
	\end{subfigure}%
	\caption{(a) sample waveform for training. (b) corresponding magnitude FT $|\hat{Y}_{m}(\omega)|$.  Spectral density of learnt kernel using (c-top) TM (red dashed line), (c-middle) ML (red), (c-bottom) FL (red).  }
	\label{fig:train_sample_and_inference}
\end{figure}

A single note produced by a music instrument (see Fig. \ref{fig:waveform}) consist of a full spectrum of harmonics with an attack and  decay in intensity. The spectrum evolution is instrument dependent, and therefore must be learnt in a recording-specific way \cite{Kirkpatrick14, Cheng16}. This motivates the design of what we call the  \textit{Mat\'{e}rn spectral mixture} (MSM) kernel; a stationary covariance function able to reflect the complex harmonic content of sounds of single notes.
%
%
%CC transition 
In this section, we first recall the spectral representation of stationary kernels. Next, we introduce the formulation of the MSM kernel by an illustrative example. 
This covariance describes the components $\left\lbrace w_{m}(t)\right\rbrace_{m=1}^{M}$.

%CC Method
%CC spectral rep of stationary kernels
Stationary GPs are those with a kernel that can be written as a function of the distance between observations \cite{rasmussen05}, that is $k(t,t') = k(|t-t'|) = k(r)$. 
The Wiener-Khintchine theorem  defines the duality of spectral densities $s(\omega)$ and stationary covariance functions $k(r)$ \cite{Shanmugan88}, specifying the following relations 
\begin{align}\label{e:WK_1}
s(\omega) = \mathcal{F} \left\lbrace k(r) \right\rbrace {\color{white} 10} &= \int_{-\infty}^{\infty} k(r) e^{-i\omega r} \dif r,
\\
\label{e:WK_2}
k(r)  = \mathcal{F}^{-1} \left\lbrace s(\omega) \right\rbrace  &= \frac{1}{2\pi}\int_{-\infty}^{\infty} s(\omega) e^{i \omega r} \dif \omega.
\end{align}
%
%Notice that $\omega$ is different to $w_m(t)$. 
Since kernels are symmetric real functions, the spectrum of the process is also a symmetric real function \cite{Hensman16}. Taking as example two basic kernels we use later on, we apply \eqref{e:WK_1} on the Mat\'{e}rn-$\frac{1}{2}$ and Cosine kernels \cite{rasmussen05}, defined as
\begin{alignat}{3}
\label{e:Matern_ker}
& k_{\sfrac{1}{2}}(r) &&= \sigma^2 e^{-\lambda r}, & \qquad \lambda &= l^{-1},\\
\label{e:cos_ker}
& k_{\text{COS}}(r)   &&= \cos({\omega_0} r),      & \qquad \omega_0 &= 2 \pi f_0,
\end{alignat}
respectively. In \eqref{e:Matern_ker} $l$ governs the time length-scale over which the function varies, and $\sigma^2$ defines the vertical variation. In \eqref{e:cos_ker} $f_0$ defines the function's frequency in Hertz, and the variance is assumed to be one. The corresponding spectral densities are

\begin{alignat}{3}\label{e:S_matern}
&s_{\sfrac{1}{2}}(\omega) &&= 2 \sigma^2 \lambda (\lambda^2 + \omega^2)^{-1},
\\
\label{e:S_cos}
&s_{\text{COS}}(\omega)  &&=  \pi
%\sqrt{\sfrac{\pi}{2}} 
\left[ 
\delta(\omega - \omega_0) + 
\delta(\omega + \omega_0)
\right].
\end{alignat}
We use the spectral representation of covariance functions to formulate the MSM kernel. Fig. \ref{fig:waveform} shows the waveform of a single note $\hat{y}_{m}(t)$, corresponding to playing pitch C4 (261.6 Hz) on an electric guitar. Fig. \ref{fig:FT_waveform} depicts the corresponding magnitude Fourier Transform (FT) $|\hat{Y}_{m}(\omega)|$, which is a real, symmetric function, similar to kernels and its corresponding spectral densities.   
This leads to the idea of designing kernels whose spectral density is close to the frequency content of the single notes available for training, that is 
%
%$
\begin{align}
s(\omega) \approx |\hat{Y}_{m}(\omega)|.
\end{align}
%$
%
However, the Mat\'{e}rn-$\frac{1}{2}$ covariance function \eqref{e:Matern_ker} is not appropriate for modelling harmonic content by itself. This is because the spectral density of this kernel has the form of a Lorentzian function (see \eqref{e:S_matern}) centred on the origin \cite{Kanevsky09}, whereas the spectral density of single notes have peaks at certain frequencies not necessarily at $\omega = 0$ (see Fig. \ref{fig:FT_waveform}).
To describe a single partial in Fig. \ref{fig:FT_waveform} it is necessary to shift the spectral density  of the Mat\'{e}rn-$\frac{1}{2}$, centring it around a specific frequency.  
To do so, we multiply \eqref{e:Matern_ker} by \eqref{e:cos_ker}, ending up with the base kernel $k(r) = k_{\sfrac{1}{2}}(r) \cdot k_{\text{COS}}(r) $. Replacing $k(r)$ in \eqref{e:WK_1}, and using the convolution theorem,  then
\begin{alignat}{2}
&s(\omega)
&&=
%\mathcal{F}\left\lbrace k(\tau) \right\rbrace
%\\ \notag
%&=
%=
%\mathcal{F}\left\lbrace \hat{k}(r) \cdot \check{k}(r) \right\rbrace 
%\\ \notag
%&= 
%=
%\mathcal{F}\left\lbrace k_{\sfrac{1}{2}}(r)  \right\rbrace
%\ast
%\mathcal{F}\left\lbrace k_{\text{COS}}(r) \right\rbrace
%=
L(\omega; \boldsymbol{\theta}) + L(-\omega; \boldsymbol{\theta}),
%\\ \notag
%&= 
%\left(
%\sigma^2 \frac{\lambda}{\lambda^2 + \omega^2} 
%\right) 
%\ast
%\left[ \delta(\omega - \omega_0) + \delta(\omega + \omega_0) \right]
%\\ \notag
%&= 
% L(\omega) + L(-\omega).
\\ \label{e:L}
&L_{}(\omega; \boldsymbol{\theta}) &&= \frac{2 \pi \sigma^2 \lambda_{} }{ \lambda_{}^{2} + (\omega - \omega_{0})^{2} },
\end{alignat}
%
%where to keep notation uncluttered we define the function $L$ as
%
%\begin{align}\label{e:L}
%L_{}(\omega; \boldsymbol{\theta}) = \frac{2 \pi \sigma^2 \lambda_{} }{ \lambda_{}^{2} + (\omega - %\omega_{0})^{2}   },
%\end{align}
%
with $\boldsymbol{\theta} = \left\lbrace \sigma^2, \lambda,  \omega_0 \right\rbrace  $, i.e. the set of hyperparameters associated with \eqref{e:Matern_ker} and \eqref{e:cos_ker}.
Expression \eqref{e:L} corresponds to shift, from the origin to $\omega_0$, the Mat\'{e}rn-$\frac{1}{2}$ spectral density \eqref{e:S_matern}.
To model $N_h$ number of partials we use a linear combination of Lorentzian functions pairs
\begin{align}\label{e:L_mix}
s_{\text{MSM}}(\omega; \boldsymbol{\Theta}) = 
\sum_{j = 1}^{N_h}  L_{}(\omega;\boldsymbol{\theta}_j) + L_{}(-\omega;\boldsymbol{\theta}_j),
\end{align}
where $\boldsymbol{\Theta} = \left\lbrace \boldsymbol{\theta}_j \right\rbrace_{j = 1}^{Nh}  $. 
%containing the parameters for each $L_{j}(\omega; \theta_j)$.
The aim of learning stage is then to find the $\boldsymbol{\Theta}$ that makes $s_{\text{MSM}}(\omega)$ close to $|\hat{Y}_{m}(\omega)|$, that is
\begin{align}\label{e:objective_fun}
\Theta^{*} =  
\underset{\Theta }{\argmin }\
%\frac{1}{N}
\sqrt{
\left( 
s_{\text{MSM}}(\omega; \boldsymbol{\Theta})  -  |\hat{Y}_{m}(\omega)|
\right)^{2}
}.
\end{align}
%
%where $N$ is the number of data samples for the single notes. 
An algorithm for optimizing \eqref{e:objective_fun} is proposed in section \ref{sec:inference}. 
Finally, replacing \eqref{e:L_mix} in \eqref{e:WK_2} we end up with a kernel with form
\begin{align}\label{equ:maternSM}
k_{\text{MSM}}(r) =   
\sum_{j=1}^{N_h} 
\sigma_j^2 
e^{-\lambda_j r} 
\cos({\omega_0}_j r ),
\end{align}
where ${\omega_0}_j$ is the frequency in radians, $\sigma_j^2$ explains the contribution of each frequency to the overall kernel, and $\lambda_j = l_j^{-1}$, where $l_j$ is the length-scale.
The MSM kernel \eqref{equ:maternSM} can be seen as a spectral-mixture kernel \cite{Wilson13}, where instead of using the squared exponential (SE) covariance we use the Mat\'{e}rn-$\frac{1}{2}$.
Although the SE kernel (a covariance function infinitely differentiable) is probably the most widely-used kernel \cite{rasmussen05}, 
in \cite{Stein19} Stein argues that such strong smoothness assumptions are unrealistic for modelling many physical processes, and recommends the Mat\'{e}rn class.
Moreover,
we have particular interest in using the family of Mat\'{e}rn kernels with half-integer orders, to explore as future work the Variational Fourier Features (VFF) recently presented in \cite{Hensman16} for efficient GP models.
Finally, by encouraging \eqref{e:L_mix} to reflect the frequency content of isolated sounds, we keep the MSM kernel within a region where it has musically-acoustically interpretation.

\subsection{Inference}
%\subsubsection{Learning kernel parameters}
\label{sec:inference}
	Learning hyperparameters by maximising the marginal likelihood is challenging because the computational complexity usually scale cubically with the number of data observations \cite{Hensman16,rasmussen05}. To overcome this, we introduce an algorithm for optimizing \eqref{e:objective_fun}. We take advantage of the sparse frequency content of the magnitude FT of the isolated events available for training (for a sample see Fig. \ref{fig:FT_waveform}). The basic idea is to fit a Lorentzian function \eqref{e:L} around each local maximum present in the spectral density, but considering only one peak at time. 
	For a step by step explanation see Algorithm \ref{algo}.
\begin{algorithm}[bh!]
	\caption{Fitting MSM kernel in frequency domain.}
	\begin{algorithmic}[1]
		\renewcommand{\algorithmicrequire}{\textbf{Input:}}
		\renewcommand{\algorithmicensure}{\textbf{Output:}}
		\REQUIRE $|\hat{Y}_{m}(\omega)|$,  $N_h$
		\ENSURE  $\boldsymbol{\Theta} = \left\lbrace \boldsymbol{\theta}_i \right\rbrace_{i = 1}^{Nh} $
		
		%\STATE $F^{*} = 1$, $c = 0$
		%\\ \textit{LOOP Process}
		\FOR {$i := 1$ \TO $N_h$}
		%\STATE $c \mathrel{+}= 1$
		\STATE $\omega^{*} = \underset{\omega}{\argmax \ } |\hat{Y}_{m}(\omega)|$
		\STATE \textit{Initialisation} $\boldsymbol{\theta} = \left\lbrace \sigma^2, \lambda,  \omega_0 = \omega^{*} \right\rbrace $
		\STATE $\boldsymbol{\theta}_i^{}  =\underset{\boldsymbol{\theta}}{\argmax \ } \sqrt{
		\left(  L(\omega;\boldsymbol{\theta}) - |\hat{Y}_{m}(\omega)|\right)^2}$  
		\STATE $|\hat{Y}_{m}(\omega)| = |L(\omega; \boldsymbol{\theta}_i) - |\hat{Y}_{m}(\omega)||$  
		\ENDFOR
		\RETURN $\boldsymbol{\Theta}$ 
	\end{algorithmic} 
	\label{algo}
\end{algorithm}

With this approach learning hyperparameters takes only few seconds, despite using all $32\times 10^3$ data points available for training for of each isolated note audio file ($16$ kHz sample frequency, $2$ seconds duration). 
Fig. \ref{fig:compare_ML_TM_FL}(top) shows  the spectral density of initializing  the MSM kernel with perfect harmonics and equal variance (dashed red line) against the actual training data frequency content (continuous blue line). Fig. \ref{fig:compare_ML_TM_FL}(middle) shows the FT of the learnt MSM kernel using marginal likelihood (red line). The frequency content of the learnt covariance using the proposed approach is depicted in Fig. \ref{fig:compare_ML_TM_FL}(bottom). 
The MSM kernel is not limited to perfect harmonics, this facilitates better fit to the audio data frequency content, which in this specific case have quasi-harmonic behaviour.

\section{Experiments}
\label{sec:results_disc}
This section presents the empirical evaluation of how \eqref{equ:factors_sum} works in practice for pitch detection.
The sigmoid (SIG) \eqref{equ:sigmoid_model} and softmax (SOF) \eqref{equ:softmax_model} model were used for inferring occurrence of two different pitches in synthetic audio of an electric guitar. In order to extend the model to more than two pitches we study the scenario where one single component $w_m(t)$ reflects the frequency content of several sound events with different pitches.  
We contrast the pitch detection performance using  an initial guess kernel or manual tuning (TM), learning in frequency domain (FL) (proposed method), and optimizing the marginal likelihood (ML).
We use the Sparse Variational GP regression implemented in \textit{GPflow} \cite{GPflow2016} for running the experiments. 
%
% CC dataset
We analysed the electric guitar audio from the study done in \cite{Yoshii13}, containing the sound events (C4, E4, G4, C4+E4, C4+G4, E4+G4, and C4+E4+G4). This signal was generated with $16$ kHz sample frequency, and last $14$ seconds. For training we used the first three isolated notes.

\subsection{Transcription of polyphonic signal}
\label{sec:amt_poly}
First we focus on detecting pitches C4 and E4, i.e. from the complete audio signal we only analysed the segments from $0$ to $4$ seconds and from $6$ to $8$ seconds. Table \ref{table:1} shows the F-measure obtained using either the sigmoid (SIG) model \eqref{equ:sigmoid_model} or the softmax (SOF) model \eqref{equ:softmax_model}. We compare how the inference approach used affects the performance of these two models. We observe that slightly better performance is achieved by using the sigmoid model. 
The learning approach considerably affects the performance of the models. The best pitch detection ($98.68\%$ F-measure) was achieved using SIG model and learning in frequency domain (FL).

\begin{figure}[th!]
	\begin{subfigure}[t]{0.5\columnwidth}
		\centering
		\includegraphics[width=1.0\columnwidth]{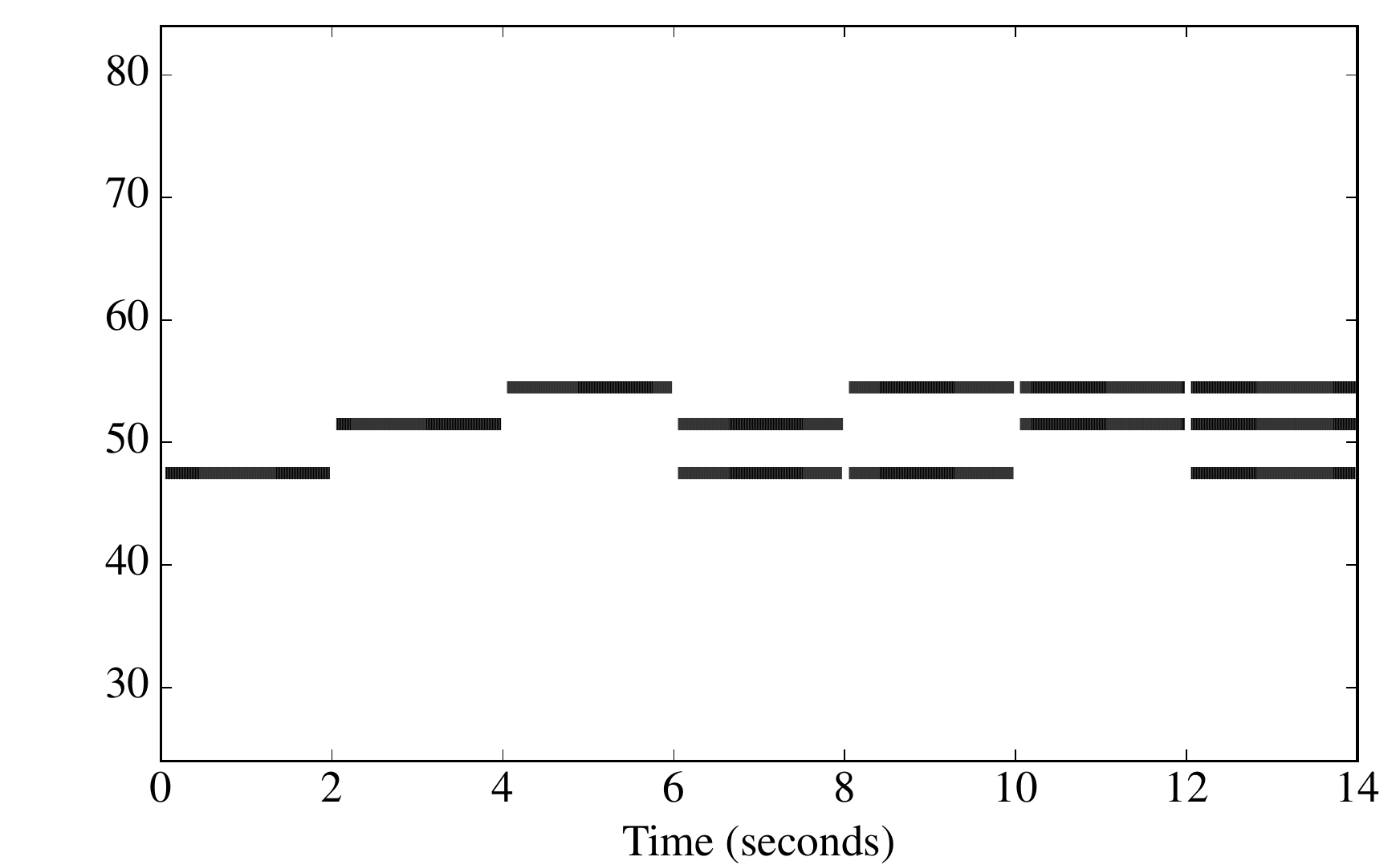}
		\caption{}
		\label{fig:ground}
	\end{subfigure}%
	%\\
	\begin{subfigure}[t]{0.5\columnwidth}
		\centering
		\includegraphics[width=1.0\columnwidth]{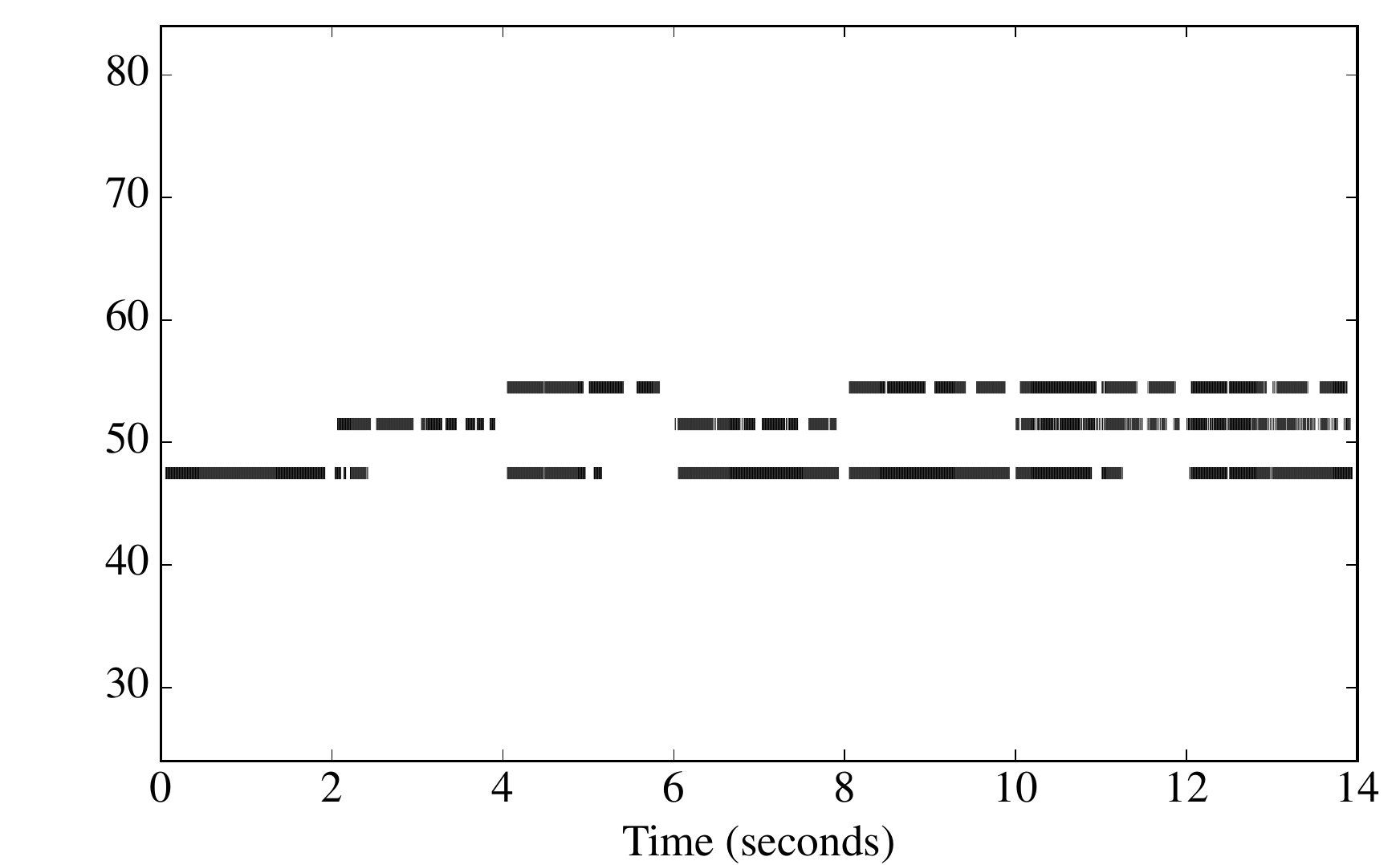}
		\caption{}
		\label{fig:T_TM}
	\end{subfigure}
	\\
	\begin{subfigure}[t]{0.5\columnwidth}
		\centering
		\includegraphics[width=1.0\columnwidth]{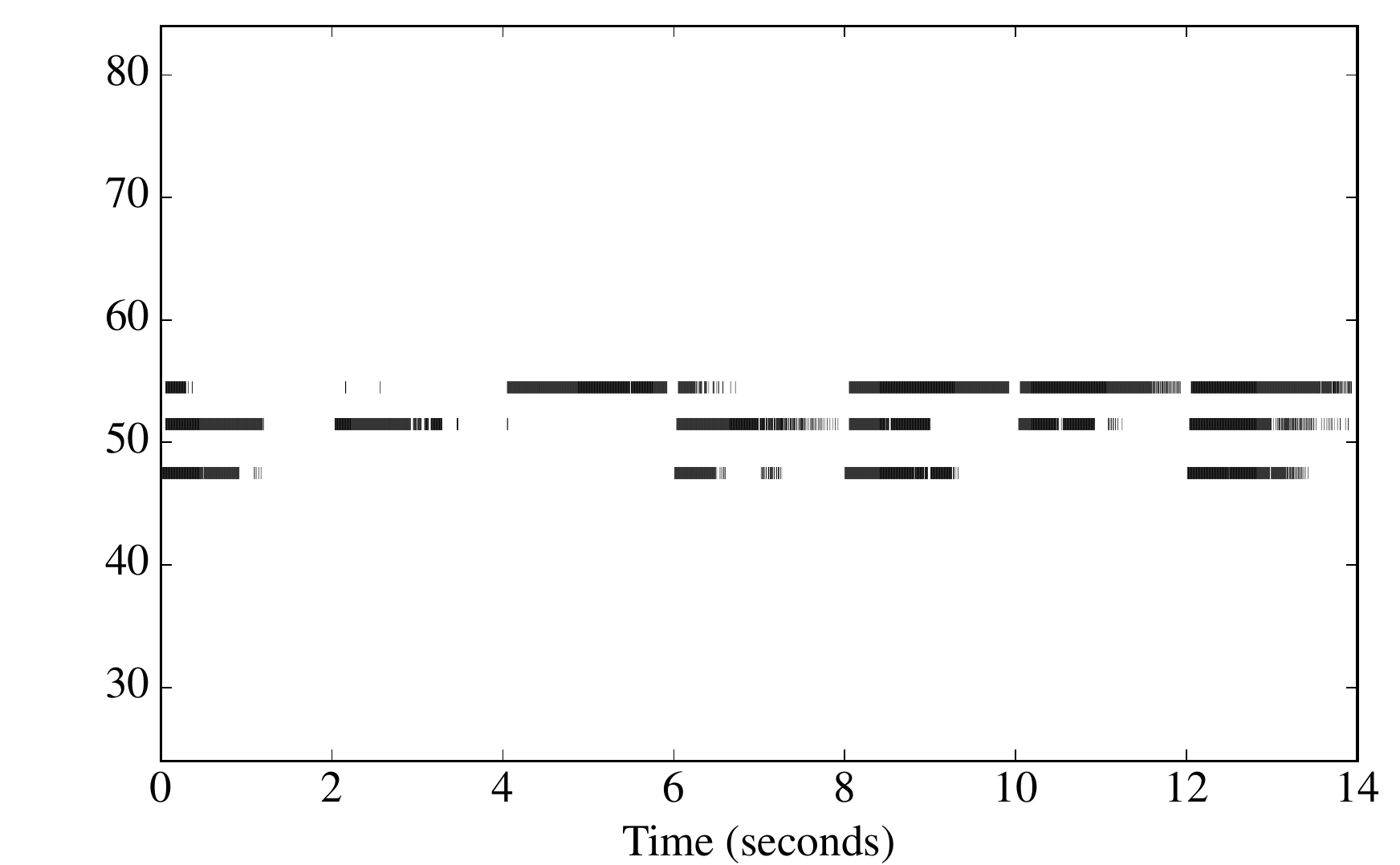}
		\caption{}
		\label{fig:T_ML}
	\end{subfigure}%
	%\\
	\begin{subfigure}[t]{0.5\columnwidth}
		\centering
		\includegraphics[width=1.0\columnwidth]{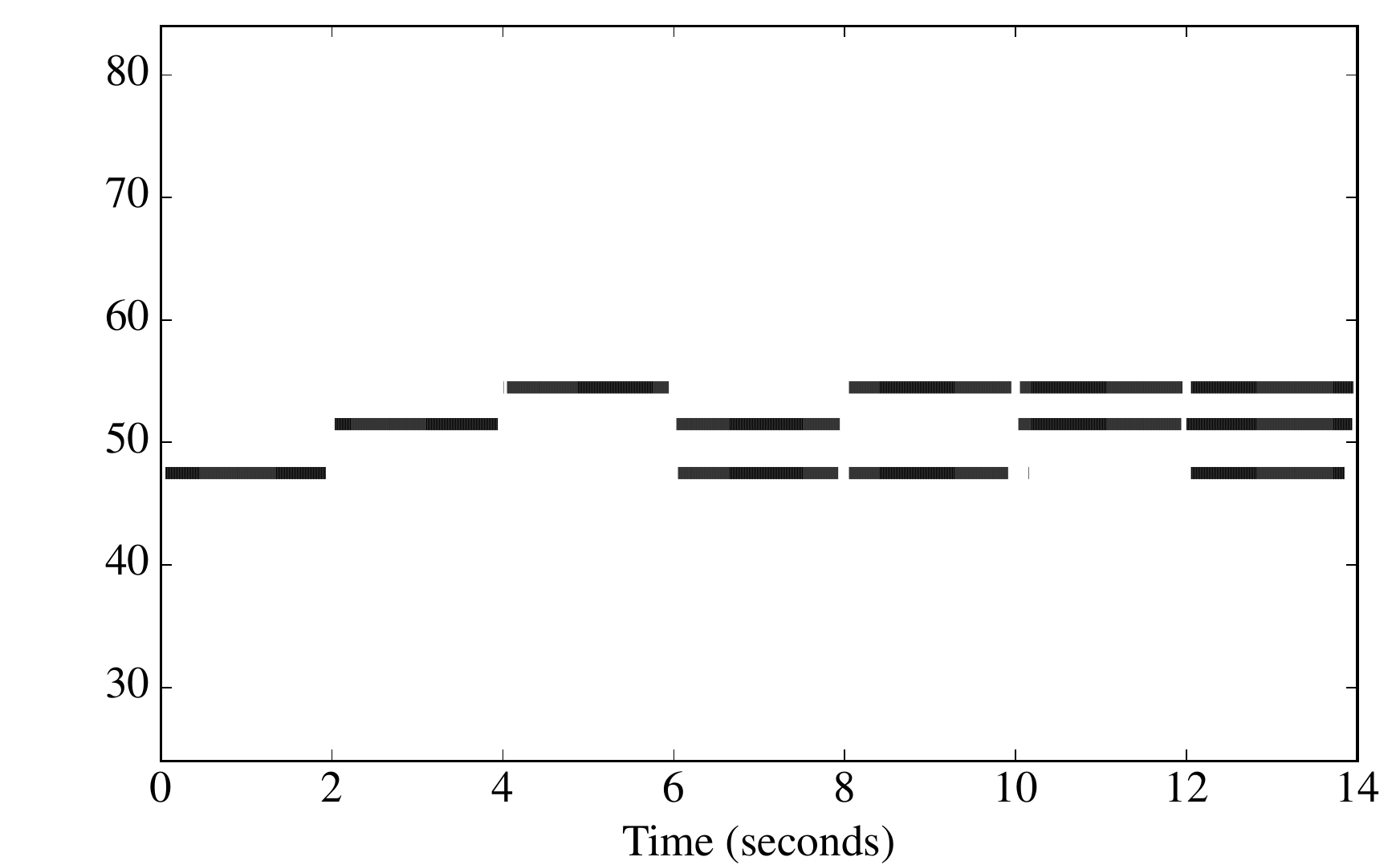}
		\caption{}
		\label{fig:T_FL}
	\end{subfigure}
\caption{Transcription using LOO-SIG. (a) ground truth. (b-d) transcription using TM, ML, FL learning approaches respectively. }
\label{fig:fig}
\end{figure}

In order to extend the model to detect more than two pitches, we allow one of the components to reflect the frequency content of two isolated notes with different pitches, per example:  $s_{1}(\omega) \approx |\hat{Y}_{\text{C4}}(\omega)| $, whereas $s_{2}(\omega) \approx |\hat{Y}_{\text{E4}}(\omega)| + |\hat{Y}_{\text{G4}}(\omega)| $. We call this approach \textit{leave one out} (SIG-LOO) as  one of the spectral densities of the covariances reflects only one pitch, whereas the other the remaining pitches. Fig. \ref{fig:ground} shows the corresponding ground truth piano-roll. Transcriptions using frequency learning, marginal likelihood optimization, and initial guess are shown in Fig. \ref{fig:T_FL}, \ref{fig:T_ML}, \ref{fig:T_TM} respectively. Results show SIG-LOO model together with the proposed learning in frequency domain outperforms for pitch detection ($98.19\%$ F-measure Table \ref{table:1}). 

\begin{table}[]
\centering
\begin{tabular}[t]{|c|c|c|c|}
	\hline
	& TM & ML & FL\\ \hline
	SIG & 89.54\%  & 59.23\% & \textbf{98.68}\% \\ \hline
	SOF & 86.28\%  & 55.28\% & 97.15\% \\ \hline\hline
	SIG-LOO & 76.21\% & 84.86\% & \textbf{98.19}\%  \\ \hline
\end{tabular}
\caption{F-measure for SIG, SOF models detecting two pitches (first two rows), and F-measure for SIG-LOO model detecting three pitches (bottom row), using three different learning approaches: TM , ML, and FL.}
\label{table:1}
\end{table}

\section{Conclusions}
\label{sec:conclusions}
We proposed a GP regression approach for pitch detection in polyphonic signals. We introduced a novel Mat\'{e}rn mixture kernel able to reflect the complex frequency content of sounds of single notes, together with an algorithm for learning its parameters in frequency domain. The proposed approach allows to introduce prior information about activations, such as smoothness (not infinite), positive-values constrains, and correlation between activations. 
Results suggest that what it is really relevant for pitch detection is a set of MSM kernels that properly fit the frequency content of the sound events to detect. Using the proposed frequency domain learning, the sigmoid model seems to be enough to perform the pitch detection, even if this models lacks to encourage dependency between activations as the softmax model does.
One advantage of using the LOO is its linear scalability regarding the number of pitches. Further empirical validation is necessary to validate its performance for more than 3 pitches.
%
% CC future work
As future work we plan to explore other Mat\'{e}rn kernels and VFF in order to be able to analyse a complete piece of music.

\section{ACKNOWLEDGMENT}
\label{sec:ack}
%{\color{white}
{\small 
James Hensman kindly shared the code for \eqref{equ:sigmoid_model} when $M = 1$.}
%}

\appendix
\section{Appendix} \label{sec:appendix}

This appendix presents the theoretical background that supports the approximate variational inference used for Gaussian process (GP) models for multi-pitch detection in time domain. 
We aim to formulate a generative probabilistic model that explains how an observed polyphonic music signal (mixture of sources) was generated. We also seek to compute a posterior distribution over the latent functions associated with each source.
The subsection \ref{chapter_single_source} covers the scenario where the mixture signal has only one source. This helps to understand the available Python code where sparse variational inference is used for making the model scalable. In subsection \ref{chapter_multi_pitch} the model is extended to several sources.

\subsection{Modulated Gaussian process} 
\label{chapter_single_source}

Here we describe the modulated GP, a model that decomposes an observed audio signal as the multiplication of a non-negative random process and a quasi-periodic Gaussian process. 
We introduce the model for one single source, defining the likelihood, prior, joint distribution, and the limitation of computing the posterior distribution.
This motivates the use of approximate inference. 
In section \ref{s:sparse_GP} we introduce inducing variables as in \cite{Hensman15}. The corresponding variational lower bound is defined in section \ref{s:lower_bound}. 
We analyse in more detail the variational expectation of the log-likelihood in section \ref{s:quadratute}, and derive two equivalent solutions, the first approximates a double integral by a two dimensional Gauss-Hermite quadrature, whereas the second solution approximates the double integral as the sum of one dimensional Gauss-Hermite quadratures. 

\subsubsection{Single source model}\label{s:single_source_model}
Given an audio recording $\mathcal{D} = \left\lbrace y_n, t_n \right\rbrace_{n=1}^{N} $
and the regression model
\begin{align*}
y(t) = \sigma(g(t))  f(t) + \epsilon(t),
\end{align*}
where $f(t)$ and $g(t)$ follow GPs and $\epsilon(t)$ follows a white noise process, then
\begin{align*}
y_n = \sigma(g_n)  f_n + \epsilon_n,
\end{align*}
where $g_n = g(t_n)$, $f_n = f(t_n)$, and $\epsilon_n \sim \mathcal{N}(\epsilon_n| 0, \nu^2)$. Defining the vectors $\textbf{y} = [y_1, \cdots, y_N]^{\top}$, $\textbf{f} = [f_1, \cdots, f_N]^{\top}$, $\textbf{g} = [g_1, \cdots, g_N]^{\top}$, and assuming the observations as i.i.d then the \textbf{likelihood} corresponds to 
\begin{align}\label{e:lik_single_source}
p(\textbf{y} | \textbf{f}, \textbf{g}) &= \prod_{n=1}^{N} p(y_n | f_n, g_n)
\\ \notag
&= \prod_{n=1}^{N} \mathcal{N}(y_n | \sigma(g_n) f_n, \nu^2).
\end{align}
We put an independent GP over each function $f(t)$ and $g(t)$, therefore the \textbf{prior} over the latent vectors $\textbf{f}$ and $\textbf{g}$ corresponds to $p(\textbf{f},\textbf{g}) = p(\textbf{f})p(\textbf{g})$, where
\begin{align*}
p(\textbf{f}) = \mathcal{N}(\textbf{f} | \boldsymbol{0}, \textbf{K}_{\textbf{f}}),
\end{align*}
and
\begin{align*}
p(\textbf{g}) = \mathcal{N}(\textbf{g} | \boldsymbol{0}, \textbf{K}_{\textbf{g}}).
\end{align*}
Given the prior and the likelihood we can define  the \textbf{joint distribution} as
\begin{align*}
p(\textbf{y},\textbf{f},\textbf{g}) 
&= p(\textbf{y}|\textbf{f},\textbf{g})p(\textbf{f}) p(\textbf{g})
\\ \notag
&=
\prod_{n=1}^{N} \mathcal{N}(y_n | \sigma(g_n) f_n, \nu^2)
\ \mathcal{N}(\textbf{f} | \boldsymbol{0}, \textbf{K}_{\textbf{f}})
\ \mathcal{N}(\textbf{g} | \boldsymbol{0}, \textbf{K}_{\textbf{g}}).
\end{align*}
The \textbf{posterior} can be calculated as
\begin{align}
p(\textbf{f}, \textbf{g}|\textbf{y}) = 
\frac{p(\textbf{y}| \textbf{f}, \textbf{g})p(\textbf{f})p(\textbf{g})}{p(\textbf{y})},
\end{align}
where the marginal likelihood is defined as
\begin{align}
p(\textbf{y}) 
= 
\int \int p(\textbf{y}| \textbf{f}, \textbf{g})p(\textbf{f})p(\textbf{g})
\ \text{d}\textbf{f}
\ \text{d}\textbf{g}.
\end{align}
Computing this expression is usually difficult due to $\mathcal{O}(N^3)$ complexity and non-tractability.   

\subsubsection{Introducing inducing variables}\label{s:sparse_GP}
Introducing inducing points $\textbf{Z} = \left\lbrace z_m \right\rbrace_{m=1}^{M} $ for both latent functions $f(t)$ and $g(t)$ and their corresponding inducing variables $\textbf{u}_{f} = \left\lbrace f(z_m) \right\rbrace_{m=1}^{M}$ and $\textbf{u}_{g} = \left\lbrace g(z_m) \right\rbrace_{m=1}^{M}$, then the joint distribution of all latent variables correspond to $p(\textbf{f}, \textbf{u}_{f}) = p(\textbf{f}| \textbf{u}_{f})p(\textbf{u}_{f})$, and $p(\textbf{g}, \textbf{u}_{g}) = p(\textbf{g}| \textbf{u}_{g})p(\textbf{u}_{g})$. The joint now follows
\begin{align*}
p(\textbf{y}, \textbf{f}, \textbf{g}, \textbf{u}_f, \textbf{u}_g) &= 
p(\textbf{y}| \textbf{f}, \textbf{g}, \textbf{u}_f, \textbf{u}_g)
p( \textbf{f}, \textbf{g}, \textbf{u}_f, \textbf{u}_g)
\\ \notag
&= 
p(\textbf{y}| \textbf{f}, \textbf{g})
p( \textbf{f}, \textbf{u}_f) 
p( \textbf{g}, \textbf{u}_g)
\\ \notag
&= 
\underbrace{p( \textbf{y}| \textbf{f}, \textbf{g})
	p( \textbf{f}| \textbf{u}_f) 
	p( \textbf{g}| \textbf{u}_g)
}_{p(\textbf{y}, \textbf{f}, \textbf{g} | \textbf{u}_f, \textbf{u}_g)}%
p( \textbf{u}_f) 
p( \textbf{u}_g)
\end{align*}
Then 
\begin{align*}
p( \textbf{y}| \textbf{u}_f, \textbf{u}_g) 
&= \int \int  p(\textbf{y}, \textbf{f}, \textbf{g} | \textbf{u}_f, \textbf{u}_g) \text{d}\textbf{f} \ \text{d}\textbf{g}
\\ \notag
&= 
\int \int  p(\textbf{y}| \textbf{f}, \textbf{g} , \textbf{u}_f, \textbf{u}_g) 
p(\textbf{f}, \textbf{g}| \textbf{u}_f, \textbf{u}_g)
\text{d}\textbf{f} \ \text{d}\textbf{g}
\\ \notag
&= 
\int \int  
p(\textbf{y}| \textbf{f}, \textbf{g} ) 
p(\textbf{f}| \textbf{u}_f)
p(\textbf{g}|\textbf{u}_g)
\text{d}\textbf{f} \ \text{d}\textbf{g}
\\ \notag
&= 
\mathbb{E}_{p(\textbf{g}| \textbf{u}_g)}
\left[ 
\mathbb{E}_{p(\textbf{f}| \textbf{u}_f)}
\left[ 
p(\textbf{y}| \textbf{f}, \textbf{g} ) 
\right] 
\right]
\\ \notag
&= 
\mathbb{E}_{p(\textbf{f}| \textbf{u}_f)p(\textbf{g}| \textbf{u}_g)}
\left[ 
p(\textbf{y}| \textbf{f}, \textbf{g} ) 
\right].
\end{align*}
Similar to \cite{Hensman15} we will use the following inequality to get a variational approximation 
\begin{align}\label{e:inequality}
\log p( \textbf{y}| \textbf{u}_f, \textbf{u}_g) 
\geq
\E_{p(\textbf{f}| \textbf{u}_f)p(\textbf{g}| \textbf{u}_g)}
\left[ 
\log
p(\textbf{y}| \textbf{f}, \textbf{g} ) 
\right].
\end{align}

\subsubsection{Variational lower bound}\label{s:lower_bound}
We assume the following variational distribution over all inducing variables 
\begin{align}\label{e:q_u}
q(\textbf{u}) &= q(\textbf{u}_f, \textbf{u}_g) &
\\ \notag
&= 
q(\textbf{u}_f) q(\textbf{u}_g) 
\\ \notag
&= 
\mathcal{N}(\textbf{m}_f, \textbf{S}_f)\mathcal{N}(\textbf{m}_g, \textbf{S}_g),
\end{align}
that is $q(\textbf{u}_f) = \mathcal{N}(\textbf{m}_f, \textbf{S}_f)$ and $q(\textbf{u}_g) = \mathcal{N}(\textbf{m}_g, \textbf{S}_g)$.
Using the standard variational equation 
\begin{align}\label{e:general_vi}
\log (\textbf{y}) 
&\geq 
\mathbb{E}_{q(\textbf{u})}\left[ \log p(\textbf{y}| \textbf{u}) \right] - \text{KL}\left(q(\textbf{u}) || p(\textbf{u}) \right)  
\\ \notag
&\geq 
\mathbb{E}_{q(\textbf{u}_f,\textbf{u}_g)}\left[ \log p(\textbf{y}| \textbf{u}_f, \textbf{u}_g) \right] - \text{KL}\left(q(\textbf{u}_f, \textbf{u}_g) || p(\textbf{u}_f, \textbf{u}_g) \right)
\\ \notag
&\geq 
\mathbb{E}_{q(\textbf{u}_f) q(\textbf{u}_g)}\left[ \log p(\textbf{y}| \textbf{u}_f, \textbf{u}_g) \right] - \text{KL}\left(q(\textbf{u}_f) q(\textbf{u}_g) || p(\textbf{u}_f) p(\textbf{u}_g) \right).
\end{align}

Replacing \eqref{e:inequality} in \eqref{e:general_vi} then 
\begin{align}\label{e:intermediate_elbo}
\log (\textbf{y}) 
&\geq 
\underbrace{\mathbb{E}_{q(\textbf{u}_f) q(\textbf{u}_g)}
	\left[ 
	\mathbb{E}_{p(\textbf{f}|\textbf{u}_f) p(\textbf{g}|\textbf{u}_g)}
	\left[ 
	\log p(\textbf{y}| \textbf{f}, \textbf{g})
	\right] 
	\right] 
}_{B}
- 
\underbrace{\text{KL}\left(q(\textbf{u}_f) q(\textbf{u}_g) || p(\textbf{u}_f) p(\textbf{u}_g) \right)}_{A}
.
\end{align}
\subsubsection{Analysing A in \eqref{e:intermediate_elbo}}
Analysing the KL divergence in \eqref{e:intermediate_elbo} we get
\begin{align*}
\text{KL}\left(q(\textbf{u}_f) q(\textbf{u}_g) || p(\textbf{u}_f) p(\textbf{u}_g) \right)
=
\int \int 
q(\textbf{u}_f)
q(\textbf{u}_g)
\log
\left\lbrace 
\frac{q(\textbf{u}_f)q(\textbf{u}_g)}{p(\textbf{u}_f) p(\textbf{u}_g)}
\right\rbrace 
\text{d}\textbf{u}_f 
\ \text{d}\textbf{u}_g,
\end{align*}
that is 
\begin{align*}
&= 
\int \int 
q(\textbf{u}_f)
q(\textbf{u}_g)
\left[ \log q(\textbf{u}_f) +\log q(\textbf{u}_g) -\log p(\textbf{u}_f)  -\log p(\textbf{u}_g)
\right] 
\text{d}\textbf{u}_f 
\ \text{d}\textbf{u}_g
\\ \notag
&= 
\int \int 
q(\textbf{u}_f)
q(\textbf{u}_g)
\left[ \log \left\lbrace \frac{q(\textbf{u}_f)}{p(\textbf{u}_f)} \right\rbrace  + \log \left\lbrace \frac{q(\textbf{u}_g)}{p(\textbf{u}_g)} \right\rbrace 
\right] 
\text{d}\textbf{u}_f 
\ \text{d}\textbf{u}_g
\\ \notag
&= 
\int \int 
q(\textbf{u}_f)
q(\textbf{u}_g) 
\log \left\lbrace \frac{q(\textbf{u}_f)}{p(\textbf{u}_f)} \right\rbrace 
\text{d}\textbf{u}_f 
\ \text{d}\textbf{u}_g
+
\int \int 
q(\textbf{u}_f)
q(\textbf{u}_g)
\log \left\lbrace \frac{q(\textbf{u}_g)}{p(\textbf{u}_g)} \right\rbrace  
\text{d}\textbf{u}_f 
\ \text{d}\textbf{u}_g
\\ \notag
&= 
\int  
q(\textbf{u}_f)
\log \left\lbrace \frac{q(\textbf{u}_f)}{p(\textbf{u}_f)} \right\rbrace 
\text{d}\textbf{u}_f 
+
\int 
q(\textbf{u}_g)
\log \left\lbrace \frac{q(\textbf{u}_g)}{p(\textbf{u}_g)} \right\rbrace  
\text{d}\textbf{u}_g,
\end{align*}
therefore
\begin{align}\label{e:_KL_simplification}
\text{KL}\left(q(\textbf{u}_f) q(\textbf{u}_g) || p(\textbf{u}_f) p(\textbf{u}_g) \right)
=
\text{KL}\left(q(\textbf{u}_f) || p(\textbf{u}_f) \right)
+
\text{KL}\left(q(\textbf{u}_g) || p(\textbf{u}_g) \right).
\end{align}
%%
%replacing this expression the lower bound \eqref{e:general_vi} we get
%%
%\begin{align*}
%\log (\textbf{y}) 
%\geq 
%\mathbb{E}_{q(\textbf{u}_f) q(\textbf{u}_g)}\left[ \log p(\textbf{y}| \textbf{u}_f, \textbf{u}_g) \right] - 
%\text{KL}\left(q(\textbf{u}_f) || p(\textbf{u}_f) \right)
%-
%\text{KL}\left(q(\textbf{u}_g) || p(\textbf{u}_g) \right).
%\end{align*}
%%
%

\subsubsection{Analysing B in \eqref{e:intermediate_elbo}}
Analysing the expectation we have
\begin{align*}
&\mathbb{E}_{q(\textbf{u}_f) q(\textbf{u}_g)}
\left[ 
\mathbb{E}_{p(\textbf{f}|\textbf{u}_f) p(\textbf{g}|\textbf{u}_g)}
\left[ 
\log p(\textbf{y}| \textbf{f}, \textbf{g})
\right] 
\right] = 
\\ \notag
&
\int
\int
\int
\int
\log p(\textbf{y}| \textbf{f}, \textbf{g})
p(\textbf{f}|\textbf{u}_f) p(\textbf{g}|\textbf{u}_g)
q(\textbf{u}_f) q(\textbf{u}_g)
\ \text{d} \textbf{f}
\ \text{d} \textbf{g}
\ \text{d} \textbf{u}_f
\ \text{d} \textbf{u}_g
= 
\\ \notag
&
\int
\int
\log p(\textbf{y}| \textbf{f}, \textbf{g})
\left[ 	 
\int
p(\textbf{f}|\textbf{u}_f) 
q(\textbf{u}_f) 
\ \text{d} \textbf{u}_f
\right]
\cdot 
\left[
\int
p(\textbf{g}|\textbf{u}_g)
q(\textbf{u}_g)
\ \text{d} \textbf{u}_g
\right] 
\ \text{d} \textbf{f}
\ \text{d} \textbf{g}
= 
\\ \notag
&
\int
\int
\log p(\textbf{y}| \textbf{f}, \textbf{g})
q(\textbf{f})
q(\textbf{g})
\ \text{d} \textbf{f}
\ \text{d} \textbf{g},
\end{align*}
therefore
\begin{align}\label{e:var_exp_simplified}
\mathbb{E}_{q(\textbf{u}_f) q(\textbf{u}_g)}
\left[ 
\mathbb{E}_{p(\textbf{f}|\textbf{u}_f) p(\textbf{g}|\textbf{u}_g)}
\left[ 
\log p(\textbf{y}| \textbf{f}, \textbf{g})
\right] 
\right]
=
\mathbb{E}_{q(\textbf{f}) q(\textbf{g})} 
\left[ 
\log p(\textbf{y}| \textbf{f}, \textbf{g})
\right],
\end{align}
where
\begin{align*}
q(\textbf{f}) =  \int p(\textbf{f} | \textbf{u}_f) q(\textbf{u}_f) \text{d} \textbf{u}_f,
\\ \notag
q(\textbf{g}) =  \int p(\textbf{g} | \textbf{u}_g) q(\textbf{u}_g) \text{d} \textbf{u}_g.
\end{align*}

\subsubsection{ELBO}
The evidence lower bound (ELBO) is defined by replacing \eqref{e:_KL_simplification} and \eqref{e:var_exp_simplified} into \eqref{e:intermediate_elbo}
\begin{align}\label{e:ELBO}
&\text{ELBO}(q(\textbf{u}_f), q(\textbf{u}_g)) = 
\\ \notag
&\mathbb{E}_{q(\textbf{f}) q(\textbf{g})} 
\left[ 
\log p(\textbf{y}| \textbf{f}, \textbf{g})
\right] - 
\text{KL}\left(q(\textbf{u}_f) || p(\textbf{u}_f) \right)
-
\text{KL}\left(q(\textbf{u}_g) || p(\textbf{u}_g) \right).
\end{align}
This is the functional we aim to maximise in the variational approach.

\subsection{Approximating variational expectations using quadrature} \label{s:quadratute}
Analysing the expectation in the lower bound equation \eqref{e:ELBO}, and using the definition of the likelihood \eqref{e:lik_single_source} we have
\begin{align}\label{e:exp_sum_quad}
\mathbb{E}_{q(\textbf{f}) q(\textbf{g})} 
\left[ 
\log p(\textbf{y}| \textbf{f}, \textbf{g})
\right] 
&= 
\mathbb{E}_{q(\textbf{f}) q(\textbf{g})} 
\left[ 
\log \prod_{n=1}^{N} p(y_n| f_n, g_n)
\right]
\\ \notag
&= 
\mathbb{E}_{q(\textbf{f}) q(\textbf{g})} 
\left[ 
\sum_{n=1}^{N} 
\log
p(y_n| f_n, g_n)
\right]
\\ \notag
&= 
\sum_{n=1}^{N} 
\mathbb{E}_{q(\textbf{f}) q(\textbf{g})} 
\left[ 
\log
p(y_n| f_n, g_n)
\right]
\\ \notag
&= 
\sum_{n=1}^{N} 
\int
\int
\log
p(y_n| f_n, g_n)
q(\textbf{f}) q(\textbf{g})
\ \text{d}\textbf{f} \ \text{d}\textbf{g}
\\ \notag
&= 
\sum_{n=1}^{N} 
\int
\int
\log
p(y_n| f_n, g_n)
q(f_n) q(g_n)
\ \text{d}f_n \ \text{d}g_n,
\end{align}
then we end up solving $N$ two dimensional Gauss-Hermite quadratures.

From \eqref{e:exp_sum_quad} we aim to approximate the following double integral by quadrature methods
\begin{align}\label{e:double_expectation}
\int
\int
\log
p(y_n| f_n, g_n)
q(f_n) q(g_n)
\ \text{d}f_n \ \text{d}g_n.
\end{align}
In the next two subsections we present two different solutions. The first one solves the double integral in \eqref{e:exp_sum_quad} by using a two dimensional quadrature, whereas the second solves \eqref{e:exp_sum_quad} by using a linear combination of 2 one dimensional quadratures, this might help to reduce computational cost. 

\subsubsection{Solving \eqref{e:double_expectation} by using quadrature of dimension 2}
From the definition of the variational distribution $q(\textbf{u}_f, \textbf{u}_g)$ in \eqref{e:q_u} we know that $q(f_n) = \mathcal{N}(f_n| m_{f_n}, s_{f_n}^2)$, and  $q(g_n) = \mathcal{N}(g_n| m_{g_n}, s_{g_n}^2)$
Then, replacing in \eqref{e:double_expectation} we get
\begin{align}\label{e:solving_2D_quadrature}
&\int
\int
\log
p(y_n| f_n, g_n)
\mathcal{N}(f_n| m_{f_n}, s_{f_n}^2)
\mathcal{N}(g_n| m_{g_n}, s_{g_n}^2)
\ \text{d}f_n \ \text{d}g_n
=
\\ \notag
&
\frac{1}{(2 \pi s_{f_n}^2)^{1/2}}
\frac{1}{(2 \pi s_{g_n}^2)^{1/2}}
\int
\int
\log
p(y_n| f_n, g_n)
\times \cdots
\\ \notag
&
\exp\left\lbrace - \frac{1}{2  s_{f_n}^2 } (f_n - m_{f_n})^2 \right\rbrace 
\exp\left\lbrace - \frac{1}{2  s_{g_n}^2 } (g_n - m_{g_n})^2 \right\rbrace 
\ \text{d}f_n \ \text{d}g_n, 
\end{align}
introducing the following change of variable: 
\begin{align*}
\hat{f}_n = \frac{f_n - m_{f_n}}{\sqrt{2} s_{f_n}}, 
\end{align*}
and
\begin{align*}
\hat{g}_n = \frac{g_n - m_{g_n}}{\sqrt{2} s_{g_n}}, 
\end{align*}

then \eqref{e:solving_2D_quadrature} can be written as 
\begin{align*}
&\int
\int
\log
p(y_n| f_n, g_n)
\mathcal{N}(f_n| m_{f_n}, s_{f_n}^2)
\mathcal{N}(g_n| m_{g_n}, s_{g_n}^2)
\ \text{d}f_n \ \text{d}g_n
=
\\ \notag
&
\frac{1}{\pi}
\int
\int
\log p(y_n| \sqrt{2} s_{f_n}\hat{f}_n + m_{f_n}, \sqrt{2}s_{g_n}\hat{g}_n + m_{g_n})
\exp\left\lbrace - \hat{f_n}^2 \right\rbrace  
\exp\left\lbrace - \hat{g_n}^2 \right\rbrace 
\ \text{d}\hat{f}_n \ \text{d}\hat{g}_n, 
\end{align*}
The previous double integral can be approximated as
\begin{align}\label{e:approx_2D}
&\int
\int
\log
p(y_n| f_n, g_n)
q(f_n) q(g_n)
\ \text{d}f_n \ \text{d}g_n
\approx
\\ \notag
&\frac{1}{\pi}
\sum_{\forall i}
\sum_{\forall j}
w_i 
w_j
\log p(y_n| \sqrt{2} s_{f_n}\hat{x}_i + m_{f_n}, \sqrt{2}s_{g_n}\hat{y}_j + m_{g_n}),
\end{align}
where $w_i, w_j, \hat{x}_i, \hat{y}_i$ are obtained from the formula for the Gauss-Hermite quadrature.

\subsubsection{Solving \eqref{e:double_expectation} by using quadratures of dimension 1}
Focusing on the expression for the likelihood of a single point $y_n$ in \eqref{e:double_expectation} 
\begin{align*}
p(y_n| f_n, g_n)
&= 
\mathcal{N}(y_n| \sigma(g_n)f_n, \nu^2)
\\ \notag
&=
\frac{1}{(2 \pi \nu^2)^{1/2}}
\exp
\left\lbrace 
-
\frac{1}{2\nu^2}
\left[y_n -  \sigma(g_n)f_n \right] ^2
\right\rbrace,
\end{align*}
then
\begin{align*}
\log p(y_n| f_n, g_n)
=
- \frac{1}{2} \log(2 \pi) 
- \frac{1}{2} \log(\nu^2)
- \frac{1}{2\nu^2} \left[y_n -  \sigma(g_n)f_n \right] ^2,
\end{align*}
replacing this into \eqref{e:double_expectation} we get
\begin{align*}
&
\int
\int
\log
p(y_n| f_n, g_n)
q(f_n) q(g_n)
\ \text{d}f_n \ \text{d}g_n 
=
\\ \notag
& 
-\frac{1}{2\nu^2}
\int
\int
\left[y_n -  \sigma(g_n)f_n \right] ^2
q(f_n) q(g_n)
\ \text{d}f_n \ \dif g_n 
- \frac{1}{2} \log(2 \pi) 
- \frac{1}{2} \log(\nu^2),
\end{align*}
where
\begin{align*}
&\int
\int
\left[y_n -  \sigma(g_n)f_n \right] ^2
q(f_n) q(g_n)
\ \text{d}f_n \ \text{d}g_n 
=
\\ \notag
&
\int
\int
\left[
y_n^2 -  2 y_n \sigma(g_n)f_n + \sigma(g_n)^2f_n^2
\right] 
q(f_n) q(g_n)
\ \text{d}f_n \ \text{d}g_n 
=
\\ \notag
&
\int
\int
y_n^2
q(f_n) q(g_n)
\ \text{d}f_n \ \text{d}g_n 
- \cdots
\\ \notag
&
\int
\int
2 y_n 
\sigma(g_n)f_n 
q(f_n) q(g_n) 
\ \text{d}f_n \ \text{d}g_n 
+ \cdots
\\ \notag
&
\int
\int
\sigma(g_n)^2f_n^2
q(f_n) q(g_n)
\ \text{d}f_n \ \text{d}g_n =
\\ \notag
&
y_n^2
-
2 y_n 
\int
f_n 
q(f_n) 
\ \text{d}f_n 
\cdot
\int
\sigma(g_n)
q(g_n)
\ \text{d}g_n 
+ 
\int
f_n^2
q(f_n) 
\ \text{d}f_n 
\cdot
\int
\sigma(g_n)^2
q(g_n)
\ \text{d}g_n =
\\ \notag
&
y_n^2
-
2 y_n 
\mathbb{E}_{q(f_n)}
\left[ 
f_n
\right] 
\mathbb{E}_{q(g_n)}
\left[ 
\sigma(g_n)
\right] 
+ 
\mathbb{E}_{q(f_n)}
\left[ 
f_n^2
\right] 
\mathbb{E}_{q(g_n)}
\left[ 
\sigma(g_n)^2
\right] =
\\ \notag
&
y_n^2
-
2 y_n 
m_{f,n}
\mathbb{E}_{q(g_n)}
\left[ 
\sigma(g_n)
\right] 
+ 
(
s_{f,n}^{2} 
+ 
m_{f,n}^{2}
)
\mathbb{E}_{q(g_n)}
\left[ 
\sigma(g_n)^2
\right].
\end{align*}
Then
\begin{align}\label{e:approx_1D}
&
\int
\int
\log
p(y_n| f_n, g_n)
q(f_n) q(g_n)
\ \text{d}f_n \ \text{d}g_n
=
\\ \notag
&
-\frac{1}{2\nu^2}
\left\lbrace 
y_n^2
-
2 y_n 
m_{f,n}
\mathbb{E}_{q(g_n)}
\left[ 
\sigma(g_n)
\right] 
+ 
(
s_{f,n}^{2} 
+ 
m_{f,n}^{2}
)
\mathbb{E}_{q(g_n)}
\left[ 
\sigma(g_n)^2
\right]
\right\rbrace 
- \frac{1}{2} \log(2 \pi) 
- \frac{1}{2} \log(\nu^2).
\end{align}
where $m_{f,n}$ and $s_{f,n}^2$ are the mean and variance of the variational distribution over the latent variable $f_n$. The expectations in the previous expression can be approximated using 2 one dimensional Gauss-Hermite quadrature. Therefore we have reduced the dimensionality of the approximate integrals.

\subsubsection{Approximating $\mathbb{E}_{q(g_n)}\left[ \sigma(g_n)\right]$ and $\mathbb{E}_{q(g_n)}\left[ \sigma(g_n)^2 \right]$ in \eqref{e:approx_1D} using 1 dimensional Gauss-Hermite quadrature}
Here we study in more detail the expectations found \eqref{e:approx_1D}. Specifically $\mathbb{E}_{q(g_n)}\left[\sigma(g_n)\right]$ and $\mathbb{E}_{q(g_n)}\left[ \sigma(g_n)^2\right]$
\begin{align*}
\mathbb{E}_{q(g_n)}\left[\sigma(g_n)\right]
=
\int_{-\infty}^{\infty}
\sigma(g_n) 
\frac{1}{(2 \pi s_{g_n}^2)^{1/2}}
\exp 
\left\lbrace 
-\frac{1}{2 s_{g_n}^2}
(g_n - m_{g_n})^2
\right\rbrace
\text{d} g_n,
\end{align*}
The Hermite-Gauss is defined for a normal distribution with zero mean, that is why we require a change of variable:
\begin{align*}
\tilde{x} =& \frac{g_{n} - m_{g_n}}{\sqrt{2} s_{g_n}}, 
\\\notag
\text{d} g_n =&  \sqrt{2} s_{g_n} \text{d}\tilde{x}, 
\end{align*}
then we have
\begin{align*}
\frac{1}{\sqrt{\pi}}
\int_{-\infty}^{\infty}
\sigma(\sqrt{2} s_{g_n} \tilde{x} + m_{g_n	} )  
\exp(-\tilde{x}^2)
\text{d}\tilde{x},
\end{align*}
calling $h(\tilde{x}) = \sigma(\sqrt{2} s_{g_n} \tilde{x} + m_{g_n	} )  $, then
\begin{align*}
\frac{1}{\sqrt{\pi}}
\int_{-\infty}^{\infty}
h(\tilde{x})  
\exp(-\tilde{x}^2)
\text{d}\tilde{x},
\end{align*}
Now we can approximate this integral using the Hermite-Gaussian quadrature, that is
\begin{align}\label{e:approx_exp_sigma_g}
\frac{1}{\sqrt{\pi}}
\int_{-\infty}^{\infty}
h(\tilde{x})  
\exp(-\tilde{x}^2)
\text{d}\tilde{x}
\approx
\frac{1}{\sqrt{\pi}}
\sum_{\forall j}
w_j
\sigma(\sqrt{2} s_{g_n} x_j + m_{g_n} ).
\end{align}
The expressions for $\mathbb{E}_{q(g_n)}\left[ \sigma(g_n)^2 \right]$ can be calculated similarly.
Replacing \eqref{e:approx_exp_sigma_g} into \eqref{e:approx_1D} we get
\begin{align}\label{e:approx_1D_final}
&\int
\int
\log
p(y_n| f_n, g_n)
q(f_n) q(g_n)
\ \text{d}f_n \ \text{d}g_n
\approx
\\ \notag
&
-\frac{1}{2\nu^2}
\left\lbrace 
y_n^2
-
2 y_n 
m_{f_n}
\left[ 
\frac{1}{\sqrt{\pi}}
\sum_{\forall i}
w_i
\sigma(\sqrt{2} s_{g_n} \hat{x}_i + m_{g_n} )
\right] 
\right.
+ \cdots
\\ \notag
&
\left.
(s_{f_n}^{2} 
+ 
m_{f_n}^{2}
)
\left[ 
\frac{1}{\sqrt{\pi}}
\sum_{\forall j}
w_j
\sigma(\sqrt{2} s_{g_n} \hat{y}_j + m_{g_n} )^2
\right]
\right\rbrace 
- \frac{1}{2} \log(2 \pi) 
- \frac{1}{2} \log(\nu^2).
\end{align}
% 

%\section{Summary}
%Then, putting all together:
%%
%\begin{align*}
%\text{ELBO}(q(\textbf{u}_f), q(\textbf{u}_g)) = 
%\mathbb{E}_{q(\textbf{f}) q(\textbf{g})} 
%\left[ 
%\log p(\textbf{y}| \textbf{f}, \textbf{g})
%\right] - 
%\text{KL}\left(q(\textbf{u}_f) || p(\textbf{u}_f) \right)
%-
%\text{KL}\left(q(\textbf{u}_g) || p(\textbf{u}_g) \right),
%\end{align*}
%where
%%
%\begin{align*}
%q(\textbf{f}) =  \int p(\textbf{f} | \textbf{u}_f) q(\textbf{u}_f) \text{d} \textbf{u}_f,
%\\ \notag
%q(\textbf{g}) =  \int p(\textbf{g} | \textbf{u}_g) q(\textbf{u}_g) \text{d} \textbf{u}_g,
%\end{align*}
%%
%and
%\begin{align*}
%&\mathbb{E}_{q(\textbf{f}) q(\textbf{g})} 
%\left[ 
%\log p(\textbf{y}| \textbf{f}, \textbf{g})
%\right] =
%\\ \notag
%&
%- \frac{N}{2} \log(2 \pi) 
%- \frac{N}{2} \log(\nu^2)
%-\sum_{n=1}^{N}
%y_n^2
%-
%2 y_n 
%m_{f,n}
%\mathbb{E}_{q(g_n)}
%\left[ 
%\sigma(g_n)
%\right] 
%+ 
%(
%s_{f,n}^{2} 
%- 
%m_{f,n}^{2}
%)
%\mathbb{E}_{q(g_n)}
%\left[ 
%\sigma(g_n)^2
%\right].
%\end{align*}
%
%Substituting tal en tal then  
%\begin{align*}
%&\mathbb{E}_{q(\textbf{f}) q(\textbf{g})} 
%\left[ 
%\log p(\textbf{y}| \textbf{f}, \textbf{g})
%\right] =
%\\ \notag
%&
%- \frac{N}{2} \log(2 \pi) 
%- \frac{N}{2} \log(\nu^2)
%- \cdots
%\\ \notag
%&
%\sum_{n=1}^{N}
%y_n^2
%-
%2 y_n 
%m_{f,n}
%\mathbb{E}_{q(g_n)}
%\left[ 
%\sigma(g_n)
%\right] 
%+ 
%(
%s_{f,n}^{2} 
%- 
%m_{f,n}^{2}
%)
%\mathbb{E}_{q(g_n)}
%\left[ 
%\sigma(g_n)^2
%\right]
%\end{align*}

\subsection{Leave one out: model with two sources} \label{chapter_multi_pitch}

\subsubsection{Likelihood}
Assuming the regression model
\begin{align*}
y(t) = 
\sum_{d=1}^{D}
\sigma(g^{(d)}(t))  f^{(d)}(t) + \epsilon(t)
\end{align*}
with $D=2$, then
\begin{align*}
y_n = 
\sigma(g^{(1)}_{n})  f^{(1)}_{n} +
\sigma(g^{(2)}_{n})  f^{(2)}_{n} +
\epsilon_n
\end{align*}
assuming the observations as i.i.d then the likelihood corresponds to 
\begin{align}\label{e:lik_2sources}
p(\textbf{y} | \textbf{F}, \textbf{G}) 
&= \prod_{n=1}^{N} p(y_n | \textbf{f}_n, \textbf{g}_n)
\\ \notag
&= \prod_{n=1}^{N} \mathcal{N}(y_n | \hat{\textbf{g}}_n^{\top} \textbf{f}_n, \nu^2),
\end{align}
where the components of the matrices $[\textbf{F}]_{n,d} = f^{(d)}_{n}$, $[\textbf{G}]_{n,d} = g^{(d)}_{n}$, then each row in $\textbf{F}$ and $\textbf{G}$ is given by
the vectors $\textbf{f}_n^{\top} = [f^{(1)}_n, f^{(2)}_n]$, $\textbf{g}_n^{\top} = [g^{(1)}_n, g^{(2)}_n]$. Finally, $ \hat{\textbf{g}}_n = [\sigma(g^{(1)}_n), \ \sigma(g^{(2)}_n)]^{\top}$ represents the non-linear transformation of the envelope processes.

\subsubsection{Analysing the log-likelihood}
From \eqref{e:lik_2sources} we get the log-likelihood
\begin{align}\label{e:log_lik_2sources}
\log p(\textbf{y} | \textbf{F}, \textbf{G}) 
&= \sum_{n=1}^{N} \log p(y_n | \textbf{f}_n, \textbf{g}_n)
\\ \notag
&= \sum_{n=1}^{N} \log \mathcal{N}(y_n | \hat{\textbf{g}}_n^{\top} \textbf{f}_n, \nu^2)
\\ \notag
&= \sum_{n=1}^{N}
\left[
-\frac{1}{2}\log (2 \pi) - 
\frac{1}{2} \log(\nu^2) -
\frac{1}{2 \nu^2}
\left(
y_n -   \hat{\textbf{g}}_n^{\top} \textbf{f}_n 
\right)^2 
\right]  
\\ \notag
&= 
-\frac{N}{2}\log (2 \pi) - 
\frac{N}{2} \log(\nu^2) -
\frac{1}{2 \nu^2}
\sum_{n=1}^{N}
\left(
y_n -   \hat{\textbf{g}}_n^{\top} \textbf{f}_n 
\right)^2  
\\ \notag
&= 
-\frac{N}{2}\log (2 \pi) - 
\frac{N}{2} \log(\nu^2) -
\frac{1}{2 \nu^2}
\sum_{n=1}^{N}
\left\lbrace  
y_n -
\left[    
\sigma(g^{(1)}_n) f^{(1)}_n + 
\sigma(g^{(2)}_n) f^{(2)}_n
\right] 
\right\rbrace  
^2.
\end{align}
We are interested in calculating
\begin{align*}
\E_{q(\textbf{F}, \textbf{G})}[\log p(\textbf{y}| \textbf{F}, \textbf{G})],
\end{align*}
Then
\begin{align*}
\int
\int
\int
\int
\log p(\textbf{y}| \textbf{f}^{(1)}, \textbf{f}^{(2)}, \textbf{g}^{(1)}, \textbf{g}^{(2)})
q(\textbf{f}^{(1)})
q(\textbf{f}^{(2)})
q(\textbf{g}^{(1)})
q(\textbf{g}^{(2)})
\ \dif \textbf{f}^{(1)} 
\ \dif \textbf{f}^{(2)} 
\ \dif \textbf{g}^{(1)} 
\ \dif \textbf{g}^{(2)}.
\end{align*}
We can calculate the previous fourth integral using a 4 dimensional Gauss-Hermite quadrature. 

Now we get an expression where only 1 dimensional quadratures are required.
\begin{align*}
&\int
\int
\int
\int
\log p(\textbf{y}| \textbf{f}^{(1)}, \textbf{f}^{(2)}, \textbf{g}^{(1)}, \textbf{g}^{(2)})
q(\textbf{f}^{(1)})
q(\textbf{f}^{(2)})
q(\textbf{g}^{(1)})
q(\textbf{g}^{(2)})
\ \dif \textbf{f}^{(1)} 
\ \dif \textbf{f}^{(2)} 
\ \dif \textbf{g}^{(1)} 
\ \dif \textbf{g}^{(2)}=
\\ \notag
&-\frac{N}{2}\log (2 \pi) - 
\frac{N}{2} \log(\nu^2) -
\\ \notag
&
\frac{1}{2 \nu^2}
\sum_{n=1}^{N}
\int
\int
\int
\int
\left[  
y_n -   
\sigma(g^{(1)}_n) f^{(1)}_n - 
\sigma(g^{(2)}_n) f^{(2)}_n
\right]  
^2
\times \cdots
\\ \notag
&q(f^{(1)}_{n})
q(f^{(2)}_{n})
q(g^{(1)}_{n})
q(g^{(2)}_{n})
\ \dif f^{(1)}_{n} 
\ \dif f^{(2)}_{n} 
\ \dif g^{(1)}_{n} 
\ \dif g^{(2)}_{n}
\end{align*}
the previous expression can be decomposed into 6 quadruple-integrals
\begin{align*}
&\int
\int
\int
\int
\left\lbrace   
y_n^2 -   
2 y_n \sigma(g^{(1)}_n) f^{(1)}_n -
2 y_n \sigma(g^{(2)}_n) f^{(2)}_n +
\left[ \sigma(g^{(1)}_n) f^{(1)}_n \right]^2 +  \cdots
\right.
\\ \notag
&\left.
2\sigma(g^{(1)}_n) f^{(1)}_n
\sigma(g^{(2)}_n) f^{(2)}_n +
\left[
\sigma(g^{(2)}_n) f^{(2)}_n 
\right]^2   
\right\rbrace  
\times \cdots
\\ \notag
&q(f^{(1)}_{n})
q(f^{(2)}_{n})
q(g^{(1)}_{n})
q(g^{(2)}_{n})
\ \dif f^{(1)}_{n} 
\ \dif f^{(2)}_{n} 
\ \dif g^{(1)}_{n} 
\ \dif g^{(2)}_{n}=
\\ \notag
&
\\ \notag
&
y_n^2 - 
2 y_n 
\left\lbrace 
\tilde{m}_{n}^{f^{(1)}} \E \left[  \sigma   \left(   g^{(1)}_{n} \right)  \right] + 
\tilde{m}_{n}^{f^{(2)}} \E \left[  \sigma   \left(   g^{(2)}_{n} \right)  \right] 
\right\rbrace 
+ 
\left[ 
\left( \tilde{m}_{n}^{f^{(1)}}\right)^2  + 
\tilde{\nu}_n^{f^{(1)}}
\right] 
\E\left[  \sigma   \left( g^{(1)}_{n}\right)^2 \right] +
\cdots
\\ \notag
&
2 \tilde{m}_{n}^{f^{(1)}} \tilde{m}_{n}^{f^{(2)}} 
\E \left[  \sigma   \left(   g^{(1)}_{n} \right)  \right] 
\E \left[  \sigma   \left(   g^{(2)}_{n} \right)  \right] +
\left[ 
\left( 
\tilde{m}_{n}^{f^{(2)}}
\right) 
^{2} + \tilde{\nu}_n^{f^{(2)}}
\right] 
\E\left[  \sigma   \left( g^{(2)}_{n}\right)^2 \right].
\end{align*}
From the last expression we conclude that the 4-dimensional integral needed to compute the expectation of the log-likelihood can be calculated as a combination of four 1-dimensional integrals. This allows to use 1D-quadrature instead of 4D-quadratures, reducing computation time and memory. Rewriting we get:
\begin{align}
& \E_{q(\textbf{F:}, \textbf{G:})}[\log p(\textbf{y}| \textbf{F:}, \textbf{G:})] =
\\ \notag
& -\frac{N}{2}\log (2 \pi) - 
\frac{N}{2} \log(\nu^2) -
\frac{1}{2 \nu^2}
\sum_{n=1}^{N}
\left\lbrace 
y_n^2 - 
2 y_n 
\left[
\tilde{m}_{n}^{f^{(1)}} \E \left[  \sigma   \left(   g^{(1)}_{n} \right)  \right] + 
\tilde{m}_{n}^{f^{(2)}} \E \left[  \sigma   \left(   g^{(2)}_{n} \right)  \right] 
\right] 
+ 
\right. 
\\ \notag
&\left. 
\left[ 
\left( \tilde{m}_{n}^{f^{(1)}}\right)^2  + 
\tilde{\nu}_n^{f^{(1)}}
\right] 
\E\left[  \sigma   \left( g^{(1)}_{n}\right)^2 \right] +
2 \tilde{m}_{n}^{f^{(1)}} \tilde{m}_{n}^{f^{(2)}} 
\E \left[  \sigma   \left(   g^{(1)}_{n} \right)  \right] 
\E \left[  \sigma   \left(   g^{(2)}_{n} \right)  \right] +
\right. 
\\ \notag
&
\left. 
\left[ 
\left( 
\tilde{m}_{n}^{f^{(2)}}
\right) 
^{2} + \tilde{\nu}_n^{f^{(2)}}
\right] 
\E\left[  \sigma   \left( g^{(2)}_{n}\right)^2 \right]
\right\rbrace.
\end{align}
%

% -------------------------------------------------------------------------
% Either list references using the bibliography style file IEEEtran.bst

\bibliographystyle{IEEEtran}
\bibliography{biblio_17}

\end{document}